\documentclass[3p]{elsarticle}
\usepackage{graphicx}
\usepackage{amsmath}
\usepackage{amsfonts}
\usepackage{array}
\usepackage{booktabs}
\usepackage{multirow}
\usepackage{subfig}
\usepackage{bm}
\usepackage{algorithm}
\usepackage{algorithmic}
\usepackage{color}
\usepackage{makecell}
\usepackage{colortbl}

\begin{document}

\begin{frontmatter}

\title{ Learning Autonomous Race Driving with Action Mapping Reinforcement Learning }

\author[1]{Yuanda Wang}
\ead{wangyd@seu.edu.cn}
\author[1]{Xin Yuan}
\ead{xinyuan@seu.edu.cn}
\author[1,2]{Changyin Sun}
\ead{cysun@ahu.edu.cn}

\address[1]{School of Automation, Southeast University, Nanjing 210096, China}
\address[2]{School of Artificial Intelligence, Anhui University, Hefei 230039, China}

\begin{abstract}
Autonomous race driving poses a complex control challenge as vehicles must be operated at the edge of their handling limits to reduce lap times while respecting physical and safety constraints. This paper presents a novel reinforcement learning (RL)-based approach, incorporating the action mapping (AM) mechanism to manage state-dependent input constraints arising from limited tire-road friction. A numerical approximation method is proposed to implement AM, addressing the complex dynamics associated with the friction constraints. The AM mechanism also allows the learned driving policy to be generalized to different friction conditions.  Experimental results in our developed race simulator demonstrate that the proposed AM-RL approach achieves superior lap times and better success rates compared to the conventional RL-based approaches. The generalization capability of driving policy with AM is also validated in the experiments.
\end{abstract}

\begin{keyword}
Autonomous race driving, reinforcement learning, safety constraint, action mapping.
\end{keyword}

\end{frontmatter}

\section{Introduction}

Autonomous driving has been a hot topic in both research and industry in recent decades \cite{grigorescu2020survey, yurtsever2020survey,liu2020computing,chen2022milestones}. Various environmental perception \cite{li2020lidar,feng2020deep,fujiyoshi2019deep}, planning \cite{claussmann2020review, chen2019autonomous}, and motion control \cite{amini2020learning, li2019reinforcement} approaches for autonomous driving have been proposed. Many of them have been successfully applied to regular cars in the market. Current autonomous driving techniques have covered most driving scenarios, including highway driving, urban driving, autonomous parking, etc. In this work, we study autonomous driving in the race driving scenario where advanced driving skills are required to fully utilize the car's handling capability and minimize the lap time. In highway and urban driving scenarios, the vehicle is operated near the equilibrium point, and the dynamic can be approximated by a linear model. Differently, in race driving, the vehicle is operated near its physical limits, and thus the complex nonlinear dynamics and constraints should be considered \cite{guiggiani2014science, rajamani2011vehicle}. These factors make autonomous race driving more challenging in terms of control than other driving scenarios. 

The autonomous race driving controller should respect multiple input constraints due to physical and safety limits. The control inputs, including steering, acceleration, and deceleration have static range limits. For RL-based methods, this constraint can be easily addressed by applying a sigmoid or tanh activation function in the output layer of the policy network. Moreover, the inputs are further restricted by the limit of tire-road friction which depends on the car's instantaneous motion states. For this kind of state-dependent constraint, most existing RL-based control methods use penalty terms in the reward function, which gives punishment to the current policy when the constraint is violated. Although this reward-shaping solution is rather simple, however, the learned policy tends to be relatively conservative. To address this state-dependent input constraint in RL-based control, we introduce the action mapping (AM) mechanism \cite{yuan2022action}, which converts the actions from the direct output from the neural network to real control inputs that satisfy the state-dependent constraints with a pre-defined mapping function. This mechanism could effectively address the conservatism associated with penalty-based solutions.

In this paper, we develop a novel numerical AM-RL framework for autonomous race driving. The state-dependent control input constraint due to the limit of tire-road friction is addressed by the AM mechanism. By establishing a mapping from unconstrained network output actions to constrained control inputs, it can be guaranteed that the vehicle is always controlled within the friction limits.  However, the vehicle dynamics related to the friction limit are rather complex, and it is very difficult to find a closed-form  expression for the mapping function. Therefore, we further propose a numerical approximation method to implement AM. Then, we incorporate AM with the twin delayed deep deterministic policy gradient (TD3) algorithm \cite{fujimoto2018addressing} to train the race driving policy. For the states, we use a set of forward-observation points to indicate the curvature of the race track ahead. The reward function is specially designed to encourage the driving policy to maximize the car's velocity along the race track while avoiding driving off the track or driving in the wrong way direction. Finally, the proposed race control approach is evaluated in our developed race simulator. The race driving policy trained with TD3 and AM obtains shorter lap times and higher success rates compared to other comparative approaches. The main contributions of this paper are summarized as follows:

\begin{enumerate}
  \item The AM mechanism is introduced to RL-based autonomous race driving control problem to address the state-dependent input constraints arising from limited tire-road friction. 
  \item The AM mechanism enables the RL-based controller to better utilize the maximum tire-road friction, addressing the conservatism often associated with conventional reward-shaping solutions. The AM mechanism also allows the learned driving policy to be generalized to different friction conditions by adapting the friction constraint in the action mapping function. 
  \item A numerical approximation method is proposed to implement the AM mechanism, overcoming the challenges of dealing with complex nonlinear dynamics with constraints. This numerical method further extends the original AM mechanism, enabling it to address constraint RL problems with a more general form. 
\end{enumerate}

The remainder of this paper is organized as follows. Section \ref{sec_related} reviewed related work about autonomous racing and safe RL. In Section \ref{sec_model}, we briefly introduce the race vehicle's dynamic model and the constraint of tire friction. In Section \ref{sec_method}, the developed RL algorithm with AM mechanism for race driving is explained in detail. The simulation experiments, results, and discussions are presented in Section \ref{sec_simulation}. At last, Section \ref{sec_conclusion} concludes this paper.

\section{Related Work} \label{sec_related}

In recent decades, model predictive control (MPC) has been one of the major methods used to address the challenge of autonomous racing for both trajectory planning and motion control. In \cite{verschueren2014towards}, the time-optimal driving control task with the constraint of race track boundaries is formulated as a nonlinear model predictive control (NMPC) problem using a generalized Gauss-Newton method. However, the tire-road friction constraint in race driving dynamics is not considered. This NMPC-based method is further extended in \cite{verschueren2016time} where the lateral tire-road friction limit is described by a Pacejka model. Then, a Hessian sequential quadratic programming optimization algorithm is employed to solve the NMPC problem. Similar MPC-based methods are also presented in \cite{scheffe2022sequential, novi2019real, liniger2017real}. In those methods, the control strategies are developed based on the vehicle's dynamic model and the race track's geometrical model. To reduce the computational complexity of the optimization problems, the models are usually simplified and linearized. This could seriously affect the performance of the controller if applied to a real vehicle. To address this issue, the MPC is incorporated with some data-driven and learning-based approaches. The measurement data from real-world experiments are used to refine the dynamic model and optimize the controller. In \cite{kabzan2019learning}, a learning-based MPC approach is proposed for autonomous racing. A relatively simple nominal vehicle model is built first. Then, the model is improved by Gaussian process regression based on the measurement data. Similarly, in \cite{rosolia2019learning}, the affine time-varying prediction model is used to approximate the vehicle model. Moreover, the model predictive path integral (MPPI) control is proposed for autonomous racing in \cite{williams2017information, williams2018information}. With the help of the model learning ability, many control strategies have achieved success in real-world experiments with a variety of race cars, including a full-size electric formula race car \cite{kabzan2019learning}, a 1:10 scale RC car \cite{rosolia2019learning}, and a 1:5 scale rally race car \cite{williams2017information,williams2018information,drews2017aggressive}.

More recently, the rapid advancement of machine learning has introduced novel solutions to complex control problems. Reinforcement learning (RL), a significant type of machine learning approach, has found extensive applications in addressing continuous control problems. The RL-based control algorithm directly trains a neural network-based control policy that maximizes a reward function by online interacting with the real system or using saved trajectories. Therefore, prior knowledge of the system dynamics might not needed, and the reward function is fully flexible. Many newly proposed RL algorithms, such as deep deterministic policy gradient (DDPG) \cite{lillicrap2015continuous}, twin delayed deep deterministic policy gradient (TD3) \cite{fujimoto2018addressing}, proximal policy optimization (PPO) \cite{schulman2017proximal}, and soft actor-critic (SAC) \cite{haarnoja2018soft}, have shown notable capability in the control of complex nonlinear dynamic systems, like quadrotor helicopters and multi-joint manipulators \cite{wang2019deterministic, hwangbo2017control, gu2017deep}. 

In many physical control applications utilizing RL, the control policy must adhere to safety constraints. In the aforementioned RL algorithms, one could impose penalties on actions violating constraints to derive a safe policy. However, this approach is often inefficient and does not ensure safety throughout the training process. To address this issue, several safe RL algorithms have been proposed. In \cite{koller2018learning}, the MPC method is combined with the RL algorithm to guarantee safe exploration in training. In \cite{achiam2017constrained}, a general-purpose RL policy search algorithm named constraint policy optimization (CPO) is proposed based on the trust region method. Reward constraint policy optimization (RCPO) \cite{tessler2018reward} is a similar safe RL algorithm that uses a penalty signal to guide the policy toward a constraint-satisfying solution. There is another kind of `plug-and-play' style safe mechanism that directly works on top of the policy to correct the actions without changing the original RL algorithm, such as the safety layer (SL) technique \cite{dalal2018safe} and the AM mechanism \cite{yuan2022action}. Although SL and AM share a similar structure, the specific approaches in building the constraint model and correcting unsafe actions are significantly different. A detailed comparison and discussion of the two methods, along with experiment results, are provided in Section \ref{sec_simulation}. Moreover, in contrast to model-free RL algorithms, many safe RL algorithms or safe mechanisms require some prior knowledge of the environment, such as the dynamic model and constraint function model. If these models are not directly accessible, they can be acquired through a model identification process guided by safe rules or human demonstration.


The RL-based control methods have also been widely applied in the fields of autonomous driving and racing. In \cite{jaritz2018end}, a rally race driving policy is learned with the A3C algorithm in an end-to-end way. The image from a forward-facing camera is directly fed into the policy network without any mediated perception. Similarly, a vision-based lateral control strategy for autonomous driving on a race track is developed in \cite{li2019reinforcement}. A convolutional neural network is built to extract track features from driver-view images. Then, a DDPG-based control policy gives control commands based on the track features and the car's speed. Furthermore, a race driving agent named `GT Sophy' has achieved super-human performance in the Gran Turismo game \cite{fuchs2021super, wurman2022outracing}. The agent is trained using an improved SAC algorithm. Instead of using images, this agent's observation of the track ahead is represented by a series of points along the centerline and each edge of the track. Similarly, Remonda \textit{et al.} \cite{remonda2019formula} propose to use the look-ahead curvature to represent the upcoming shape of the track, and train policies with diverse variants of DDPG (with long short-term memory, prioritized experience replay, multi-step target, etc.). Their proposed approach outperforms the state-of-the-art bots in the TORCS simulator \cite{wymann2000torcs} and even surpasses professional drivers in qualifying sessions in a professional simulator \cite{remonda2021comparing}.

In the aforementioned RL-based autonomous racing approaches, the safety constraint is not explicitly considered. Due to the complexity of autonomous racing tasks, those general-purpose safe RL algorithms are difficult to be applied. For example, implementing the CPO algorithm in autonomous racing is quite challenging as it involves evaluating constraint functions to determine the feasibility of a certain control policy. Consequently, the safety RL methods for autonomous racing are usually specially designed. Niu \textit{et al.} \cite{niu2020two} propose a two-stage safe RL racing approach. In the first stage, a rule-based safeguard module is employed to enforce the constraint during the policy training at low speed. Then, in the second stage, the rule-based module is replaced with a data-driven module to develop a closed-form analytical safety solution at high speed. This approach is validated with the TORCS simulator and achieves zero safety violations. In \cite{evans2023safe}, a viability theory-based safety supervisory architecture is proposed. The supervisor is built with a viability kernel based on the car's dynamic model. It ensures the vehicle stays within the friction limit while maintaining recursive feasibility during the training process. Among the approaches discussed above, incorporating an additional safety module emerges as a common and logical solution for safe RL in autonomous racing. Our proposed approach, featuring the AM mechanism, also adopts this strategy.

\section{Vehicle Model} \label{sec_model}

\begin{figure}[h!]
\centering
  \includegraphics[width=0.45\linewidth]{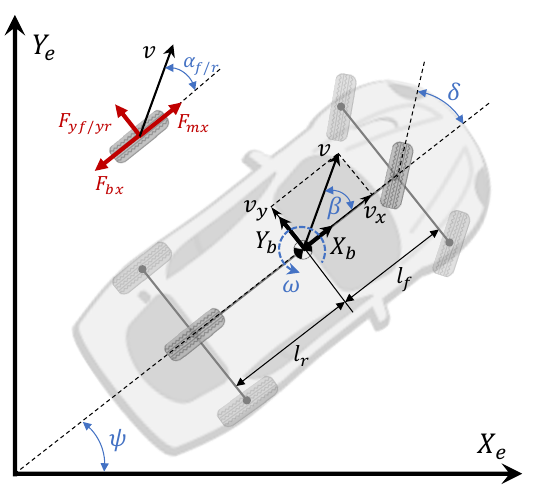}
  \caption{Single track vehicle model in earth-fixed frame and vehicle-fixed frame.}
  \label{carmodel}
\end{figure}

In this section, the single track vehicle model and friction constraint for autonomous race driving are presented. The single track model, or known as the `bicycle model', is commonly used in car handling studies \cite{guiggiani2014science}. As shown in Fig. \ref{carmodel}, based on the assumption that both front wheel steering systems have equal gear ratio, the dynamics and kinematics of two front wheels and two rear wheels can be represented by two center wheels located on the center of the front and rear axles. The pose and motion of the race car are defined under two coordinate frames: earth-fixed frame ($o_e$-$X_eY_eZ_e$) and vehicle body-fixed frame ($o_b$-$X_bY_bZ_b$).

The notations used to describe the vehicle model in Fig. \ref{carmodel} are listed as follows:

\begin{itemize}
	\item $v$: vehicle velocity at center of gravity in body-fixed frame.
	\item $v_x, v_y$: vehicle longitudinal and lateral velocity in body-fixed frame.
	\item $\psi$: vehicle heading angle (yaw angle).
	\item $\omega$: vehicle turning rate.
	\item $\delta$: front wheel steering angle.
	\item $\beta$: vehicle sideslip angle.
	\item $l_f, l_r$: distance from center of gravity to front/rear axle.
	\item $\alpha_f, \alpha_r$: tire sideslip angles of front/rear wheels.
	\item $F_{yf}, F_{yr}$: lateral tire force on front/rear wheels.
	\item $F_{mx}$: traction force on driven wheel. 
	\item $F_{bx}$: braking force on wheels.
\end{itemize}

Moreover, the effects of longitudinal and lateral load transfer are omitted. The race track is assumed to be flat and has uniform friction coefficient. In the following, the vehicle's longitudinal dynamics, lateral dynamics, and the tire friction constraint.


\subsection{Longitudinal Dynamics}
The longitudinal model describes the vehicle's motion along $X_b$-axis of the body-fixed frame. A force balance along the vehicle's longitudinal direction yields:
\begin{equation}
	m\dot{v}_x = F_{tx} - F_{aero} - F_{roll}
\end{equation}
where $m$ is the total mass of the vehicle, $F_{tx}$ is longitudinal tire force, $F_{aero}$ is aerodynamic drag force, and $F_{roll}$ is rolling resistance force. The longitudinal tire force is the friction force from the track that acts on the tire. It is the reacting force of the traction force or braking force. 

\subsubsection{Traction Force}
The traction force is generated by the vehicle's powertrain and acts on the driven wheels. Different from a gasoline engine with a gearbox, the electric motor has special characteristics when generating torques at different speeds. If the motor speed is no greater than the base speed or rated speed, that is, $n_\text{m} \leq n_{\text{b}}$, the motor works in the constant torque region, and the output torque can be directly controlled by the motor control signal. The traction force is 
\begin{equation}
	F_{mx} = \frac{K_m u_m}{R_w}
\end{equation}
where $K_m$ is the motor torque coefficient, which is determined by the transmission efficiency and the reduction gear ratio, $R_w$ is the radius of wheels, $u_m \in [0,1]$ is the normalized motor control input. If the motor speed is greater than the base speed, that is, $n_\text{m} > n_{\text{b}}$, the motor works in the constant power region. The output torque is limited by the maximum power $P_\text{max}$. The traction force is 
\begin{equation}
	F_{mx} = \min \Big[\frac{K_{\text{m}}u_m}{R_w}, ~~\frac{P_\text{max}}{n_m R_w} \Big]
\end{equation}

\subsubsection{Braking Force}
The braking force is generated by the vehicle's braking system and acts on all wheels. We assume the braking force is proportional to the braking control signal, that is, 
\begin{equation}
	F_{bx} = K_{\text{b}} u_b
\end{equation}
where $u_b \in [0,1]$ is the normalized braking control signal, $K_{\text{b}}$ is the coefficient of the braking system.

\subsubsection{Aerodynamic Drag Force}
The equivalent aerodynamic drag force on a vehicle in a windless environment can be given by
\begin{equation}
	F_{aero} = \frac{1}{2}\rho_A C_d A_f v_x
\end{equation}
where $\rho_A$ is the density of air, $C_d$ is the aerodynamic drag coefficient, $A_F$ is the frontal area of the vehicle. 

\subsubsection{Rolling Resistance Force}
The rolling resistance is roughly proportional to the down force on the tires, that is, 
\begin{equation}
	F_{roll} = f_{r} mg
\end{equation}
where $f_r$ is the rolling resistance coefficient, and $g$ is the acceleration due to gravity.

The longitudinal motion of the car is controlled by the motor control signal $u_m$ and the brake control signal $u_b$. However, two control inputs cannot be applied simultaneously in real situations. Therefore, we combine them into a single control signal $u_x \in [-1,1]$, where $u_x = 1$ denotes full motor power and $u_x = -1$ denotes full brake.


\subsection{Lateral Dynamics}
The lateral model describes the vehicle's translational motion along $Y_b$-axis and rotational motion around $Z_b$-axis. A force balance along the vehicle's lateral direction yields
\begin{equation}
	m\dot{v}_y  =F_{yf} + F_{yr}
\end{equation}
A moment balance around $Z_b$-axis yields the rotational dynamics as
\begin{equation}
	I_z \dot{\omega} = F_{yf}l_{f} - F_{yr}l_{r}
\end{equation}
where $I_z$ is vehicle body moments of inertia about the body-fixed $Z_b$-axis. The lateral tire forces $F_{yf}$ and $F_{yr}$ are proportional to their corresponding tire sideslip angles, that is
\begin{align}
	F_{yf} &= 2C_{\alpha_f} \alpha_f =2C_{\alpha_f}(\delta - \theta_{v_f}) \\
	F_{yr} &= 2C_{\alpha_r} \alpha_r =2C_{\alpha_r}(- \theta_{v_r})
\end{align} 
where $C_{\alpha_f}$ and $C_{\alpha_r}$ are cornering stiffness of the front and rear tires, $\theta_{v_f}$ and $\theta_{v_r}$ denote the direction of the velocity of front and rear wheels, and they can be obtained by
\begin{align}
	\theta_{v_f} = \arctan \frac{v_y + \omega l_f}{v_x} \\
	\theta_{v_r} = \arctan \frac{v_y - \omega l_r}{v_x}
\end{align}

Finally, the vehicle's velocity in earth-fixed frame can be expressed with the following kinematic equations:
\begin{align}
	\dot{X} &= v \cos(\psi+\beta) \\
	\dot{Y} &= v \sin(\psi+\beta) 
\end{align}

Moreover, the lateral motion of the vehicle can also be approximated by a simpler kinematic model with the assumption that the slip angles of the front and rear wheels are both zero. Then, the slip angle and turning rate is expressed by:
\begin{align}
	\beta &= \arctan\Big(\frac{l_r\tan{(\delta)}}{l_f+l_r}\Big) \\
	\dot{\psi} &= \omega = \frac{v\tan(\delta)\cos(\beta)}{l_f+l_r}
\end{align}

For an autonomous racing vehicle, the steering angle of the front wheels $\delta$ is actuated by the electric power steering system. The input can be regarded as a normalized steering angular speed signal $u_y \in {[-1,1]}$. $u_y = 1$ or $-1$ makes the power steering system actuate steering wheels to left or right at the maximum angular speed $\dot{\delta}_{\max}$, and $u_y = 0$ means maintain the current steering angle. 


\subsection{Constraint of Tire Friction}\label{subsec_constraint}
\begin{figure}[ht]
\centering
  \includegraphics[width=0.4\linewidth]{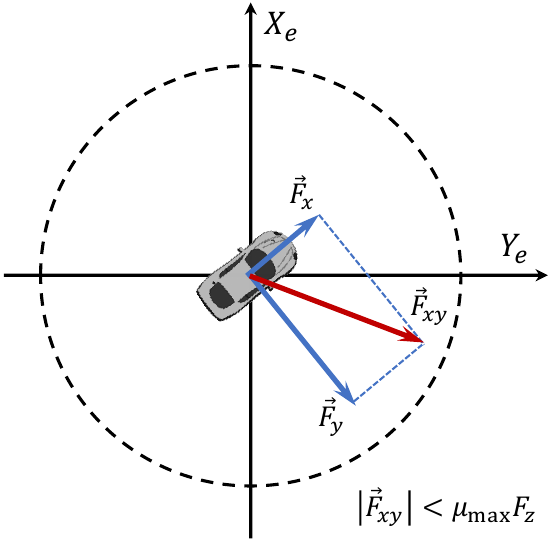}
  \caption{Constraint of tire friction described by the friction circle}
  \label{frictioncircle}
\end{figure}

The longitudinal and lateral forces that control the vehicle rely on the friction force or grip between the tire and the road. In real conditions, the friction force has an upper limit, which depends on a number of physical factors, including road surface material, tire size, tire pressure, tire temperature, etc. If we assume these physical factors remain the same during the race, we could use a friction circle model to describe the constraint of tire friction. As shown in Fig. \ref{frictioncircle}, the friction circle model indicates that the total resultant force $\vec{F}_{xy}$ applied on the vehicle cannot exceed the maximum friction force, which is 
\begin{equation}
	|\vec{F}_x + \vec{F}_y| = |\vec{F}_{xy}| < \mu_{\max}F_z
\end{equation}
where $\vec{F}_x$ and $\vec{F}_y$ are the longitudinal and lateral resultant force vectors, $F_z$ is the total vertical load on the tires. $\mu_{\text{max}}$ is the maximum tire-road friction coefficient. The aerodynamic lift force and downforce are omitted in the vehicle model, and we have $F_z = mg$. Then, the constraint can also be represented by $|\vec{a}_{xy}|<\mu_{\max}g$, where $\vec{a}_{xy}$ is the resultant acceleration measured from the vehicle. During the race, sudden and large braking or steering input, especially at high speed, could easily exceed the upper limit of grip, which could result in side slip or tail slip, and bring the vehicle into an uncontrollable spin. To guarantee the vehicle is controllable and stable in a race, we should always consider the constraint of tire friction when designing an autonomous race driving controller. 



\section{Race Driving with AM-RL} \label{sec_method}
In this section, the autonomous race driving policy is developed with an RL-based method. To guarantee the driving policy satisfies the safety constraint of tire friction in both training and implementation, we apply our proposed AM mechanism to the RL algorithm. In the following, we first introduce the basics of RL and the race driving Markov decision process (MDP) model. Then, the AM mechanism for the tire friction constraint is described. Finally, we present the implementation process of applying AM to a specific RL algorithm and train an autonomous race diving policy.

\subsection{Race Driving MDP Model}

In this subsection, the race driving MDP model for RL-based autonomous racing approaches is presented. The basic idea of RL is iteratively optimizing the control policy of an agent based on the input-output experiences from a step-by-step agent-environment interaction system. The goal is to maximize the accumulated every-step reward. The agent-environment interaction system is generally modeled as an MDP, which is defined by a tuple $\langle \mathcal{S}, \mathcal{A}, \mathcal{P}, \mathcal{R}, \gamma \rangle$, where $\mathcal{S}$ is the set of states, $\mathcal{A}$ is the set of actions, $\mathcal{P}$ and $\mathcal{R}$ are the state transition model and the reward function respectively, $\gamma \in [0,1]$ is the discount factor. The policy is denoted as $\pi$ which maps states to actions. For each step of the interaction at time step $t$, the agent in state $s_t \in \mathcal{S}$, takes action $a_t \in \mathcal{A}$ following its policy $\pi$, then receives a reward $r_{t+1}$ with the reward function $\mathcal{R}$, and finally enters a new state $s_{t+1}$ following the state transition model $\mathcal{P}$.  The accumulated reward from step $t$ onward is defined as the return, $R_t=\sum^{T_{\max}}_{t'=t}\gamma^{t'-t}r_{t'}$. The expected return following the policy $\pi$ can be represented by the state-action value function, which is defined as $Q^{\pi}(s,a)=\mathbb{E}_{\pi}[R_t|s_t,a_t]$.
 
The goal of race driving in this study is to finish a lap of the track safely in the shortest possible time. To solve the race driving task with RL, the race driving MDP model which expresses the interaction between an autonomous driving agent and a car-track environment is established first. In this MDP, the state transition model is determined by the aforementioned vehicle dynamic model. The fundamental principle of building the state and action space is to make the agent-environment interaction satisfy the Markov property to the largest extent. Therefore, all system states that could affect the race car's motion should be included. Following this principle, the state, action, and reward function are defined and described as follows. 
 
\begin{figure}[ht]
\centering
  \includegraphics[width=0.5\linewidth]{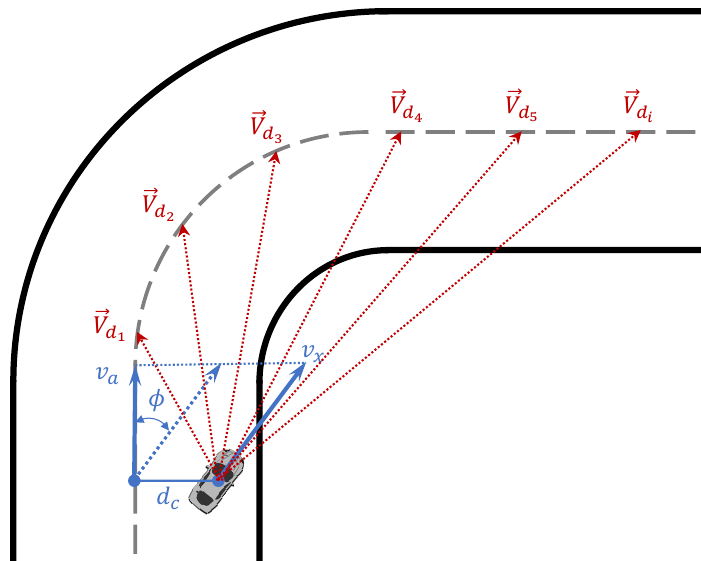}
  \caption{Part of the states defined in our race driving MDP.}
  \label{fig_track}
\end{figure}

\textbf{State:} The states of the race driving MDP are divided into two groups. The first group of states describes the race car's pose and motion. In the first group of states, $v_x$, $\omega$, and $\delta$ denote the vehicle's longitudinal velocity, turning rate, and front wheel steering angle, respectively. These states have been defined in Section \ref{sec_model}. The car's position and orientation with respect to the track are expressed by the relative cross-centerline distance $d_c \in [-1, 1]$ and the relative heading angle $\phi$. As shown in Fig. \ref{fig_track}, $d_c$ is the distance from the car's center of gravity to the centerline of the track. $\phi \in (-\pi, \pi]$ is the car's heading angle with respect to the tangential direction of the centerline projected point (orthogonal projection of the car's center of gravity onto the centerline). $\phi=0$ means the car is precisely following the track direction, while $\phi > \pi/2$ or $\phi < -\pi/2$ means the car is traveling in the opposite direction of the track, also known as wrong-way driving. The second group of states is composed of $N_{\text{FO}}$ forward-observation vectors, which are used to indicate the curvature of the race track ahead. As shown in Fig. \ref{fig_track}, the vectors $\vec{V}_{d_i}$ (red dotted arrow lines) all start from the car's center point $(X,Y)$ and point to the forward-observation points $(X_{d_i}, Y_{d_i})$ on the centerline. The forward-observation points are determined by moving the car's centerline projected point forward along the centerline over a distance of $d_i$. Specifically, a forward-observation vector $\vec{V}_{d_i}$ is defined as:
\begin{equation}
	\vec{V}_{d_i}=\mathbf{R}^{b}_{e}(\psi)[X_{d_i}-X,~~ Y_{d_i}-Y]^T
\end{equation}
where $\mathbf{R}^{b}_{e}(\psi) \in \mathbb{R}^{2 \times 2}$ is the rotational  matrix that transforms the forward-looking vector from the inertial frame to the body-fixed frame. Then, $N_{\text{FO}}$ forward-observation vectors are connected and form a forward-observation feature $\mathbf{V}_{\text{FO}} = [\vec{V}_{d_1}, \vec{V}_{d_2}, \ldots, \vec{V}_{d_{N_\text{FO}}}] \in \mathbb{R}^{2 \times N_{\text{FO}}}$. The state vector of the race driving MDP is defined as:
\begin{equation}
	s_t = [v_x, \omega, \delta, d_c, \phi, \mathbf{V}_{\text{FO}}]
\end{equation}

\textbf{Action:} The vehicle is controlled by the motor/brake control signal $u_x$ and steering control signal $u_y$. Therefore, we define two actions, $a_x$ and $a_y$, as the output from the race driving policy. The action vector is defined as:
\begin{equation}
	a_t = [a_x, a_y]
\end{equation}
The relationship between the actions and the control signals is further explained in Section \ref{AM}.



\textbf{Reward Function:}
The reward function guides the RL algorithm to optimize the policy toward a certain objective. To define a proper reward function for the autonomous race driving task, the objective of finishing a lap safely in the shortest time should be broken down into each time step. From the perspective of a single time step, the driving policy should maximize the car's velocity along the track direction, which is denoted by $v_a$ as shown in Fig. \ref{fig_track}. Therefore, our velocity reward is set as: $r_{\text{vel}} = v_a = v_x\cos{\phi}$. Furthermore, to guide the driving policy to operate safely, we set negative rewards $r_\text{out}=-100$ for driving off the track,  $r_\text{ww}=-100$ for driving in the wrong way direction, and $r_\text{tf}=-100$ for violation of tire friction constraint. If the car is safely operated inside the track, $r_{\text{out}}=r_\text{ww}=r_\text{tf}=0$. Then, the reward function is defined as:
\begin{equation}
	r_t = r_{\text{vel}} + r_{\text{out}} + r_{\text{ww}} + r_\text{tf}
\end{equation}
 It is noted that the negative reward $r_\text{tf}$ only applies to conventional RL algorithms that are not specially designed to address the friction constraint. For the proposed AM-RL approaches, $r_\text{tf}$ is not needed. 

\subsection{AM mechanism for Tire Friction Constraint}\label{AM}
 The autonomous racing control policy should generate proper control inputs that satisfy the constraint of tire friction to prevent the car from entering uncontrollable states. Based on the car's dynamic model, the friction constraint dynamically changes and it is dependent on the car's speed and steering angle. In this subsection, we proposed a numerical AM mechanism to tackle this type of state-dependent input constraint of tire friction.

The AM mechanism addresses the state-dependent constraint by establishing a mapping between an unconstrained virtual policy $\pi_\text{V}$ and a constrained real policy $\pi_\text{R}$. The virtual policy is represented by the neural network which directly maps states to actions, and it does not consider the constraints. Next, the virtual policy is converted to its corresponding constrained real policy that satisfies the state-dependent constraint. The real policy directly interacts with the real system, while the virtual policy is optimized with the RL algorithm using the interaction experiences.

More specifically, we define the unconstrained virtual policy as $a = \pi_\text{V}(s)$, where $s \in \mathcal{S}$ is the state vector. Here we define $\mathcal{S}$ as a compact subspace of $\mathbb{R}^{N_s}$, and $N_s$ is the dimension of the state space. $a \in \mathcal{A}$ is the virtual action, and $\mathcal{A}$ is the unconstrained action space. To be compatible with the neural networks represented policy, in the following, the unconstrained action space $\mathcal{A}$ is defined as $[-1,1]^{N_a}$ which is a compact subspace of $\mathbb{R}^{N_a}$, and ${N_a}$ is the dimension of the action space. Let $G$ be a compact set-valued map from $\mathcal{S}$ to the power set $P(\mathbb{R}^{N_a})$, and $G$ is characterized by its graph $\text{Graph}(G)=\{(x,y)|x \in \mathcal{S}, y \in G(x)\}$. In this work, we use $G(s)$ to denote the control input space that satisfies the state-dependent constraint. Then, the constrained real policy is defined as $u = \pi_{\text{R}}(s)$, where $u \in G(s)$ is the real control input. 

 Let $\mathcal{F}$ be the set of all continuous functions mapping from $\mathcal{S}$ to $\mathcal{A}$. Let $\mathcal{H}$ be the set of all continuous functions $\pi : \mathcal{S} \rightarrow \bigcup_{s \in \mathcal{S}} G(s)$ that satisfying the range of $\pi(s)$ in $G(s)$. We assume that the union of all graphs of map $\pi$ in $\mathcal{H}$ is equal to $\text{Graph}(G)$, that is, $\text{Graph}(G)=\bigcup_{\pi\in \mathcal{H}}\{(s,\pi(s))|s\in \mathcal{S} \}$. Then, the connection between the virtual unconstrained policy $\pi_{\text{V}} \in \mathcal{F}$  and the real constrained policy $\pi_{\text{R}} \in \mathcal{H}$ can be described as a map $T : \mathcal{H} \rightarrow \mathcal{F}$. According to the action mapping theorem (Theorem 1 in \cite{yuan2022action}), the map $T : \mathcal{H} \rightarrow \mathcal{F}$ exists if and only if there exists a continuous map $h : \text{Graph}(G) \rightarrow \mathcal{A}$ such that, for each $s \in \mathcal{S}$, the map $h_s : G(s) \rightarrow \mathcal{A}$ is the homeomorphism of $G(s)$ with $\mathcal{A}$ and $h_s$ is defined as $h_s(a) = h(s, a)$. 
 
\begin{figure}[ht]
\centering
  \includegraphics[width=0.9\linewidth]{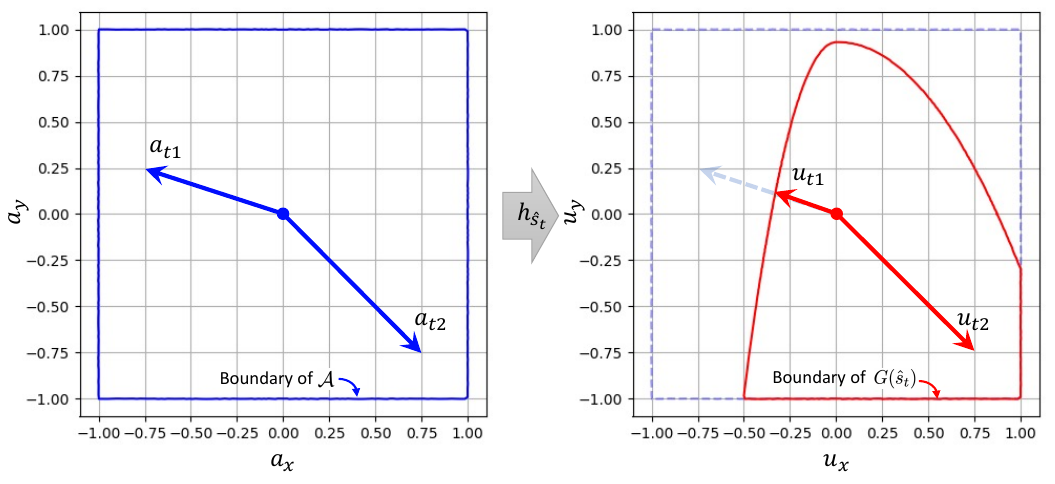}
  \caption{Action mapping example at state: $v_x=15.4~\text{m/s}, \delta=7.9~\text{deg}$. The virtual action vector examples $a_{t1}$ and $a_{t2}$ and the boundary are shown on the left. The real control input examples $u_{t1}$ and $u_{t2}$ are shown on the right.}
  \label{fig_actionmap1}
\end{figure}

 In the following, the tire friction constraint in race driving is studied under the framework of AM. The unconstrained virtual policy is constructed as $a_t=\pi_{\text{V}}(s_t)$, where $a_t=[a_x, a_y]$ is the virtual action vector, and $a_x, a_y \in [-1, 1]$ denote the virtual longitudinal action and virtual lateral action respectively. This virtual policy is represented by a neural network and optimized through RL algorithms. The constrained real policy is denoted by $u_t = \pi_{\text{R}}(s_t)$, where $u_t=[u_x, u_y]$ is the real control input vector, and $u_x$, $u_y$ are the normalized motor/brake and the steering control input respectively. The constrained real policy can be obtained by the policy space mapping $T:\mathcal{H} \rightarrow \mathcal{F}$, which is realized through its corresponding continuous mapping function:
 \begin{equation}
 	u_t = h(\hat{s}_t, a_t)
 \end{equation}
 This continuous mapping function maps the action $a_t$ from the unconstrained virtual policy to the real control input $u_t \in G(\hat{s}_t)$ that satisfies the tire friction constraint. $\hat{s}_t=[v_x, \delta]$ is a subset of the state vector which only contains the state variables related to the tire friction constraint. An illustrative example of the action mapping while the race car's longitudinal velocity $v_x=15.4~\text{m/s}$, front wheel steering angle $\delta=7.9~\text{deg}$ is given in Fig. \ref{fig_actionmap1}. The boundary of the unconstrained action space $\mathcal{A}$ is shown on the left, and the action $a_t$ can be freely selected inside the boundary. The boundary of the constrained control input space $G(\hat{s}_t)$ is shown on the right, and the shape of the boundary depends on the race car's current states $\hat{s}_t$. Any control input vector located outside the boundary will violate the tire friction constraint. In Fig. \ref{fig_actionmap1}, we give two action vector examples $a_{t1}=[-0.75, 0.25]$ and $a_{t2}=[0.75, -0.75]$ marked in blue arrows. If we directly use those two actions as the control inputs to the real system, $a_{t2}$ satisfies the constraint while $a_{t1}$ fails. Therefore, in this task, an intuitive explanation of the action mapping is to map the $a_t$ to its corresponding $u_t$ and guarantee it is inside the boundary of $G(\hat{s}_t)$. 
 
 However, due to the complexity of the vehicle dynamics and constraints, it is rather difficult to give a closed-form expression of the mapping function. Therefore, we provide a numerical approximation method to implement the action mapping mechanism for a complex dynamic system with state-dependent input constraints. The basic idea is to shorten those overrun action vectors to the boundary of $G(\hat{s})$ while keeping the same direction. For the convenience of processing the vectors, both action vectors and control input vectors are temporarily converted to polar coordinate form. The action and control inputs are expressed by $[\rho_{a}, \vartheta_{a}]$ and $[\rho_{u}, \vartheta_{u}]$ respectively. $\rho_{a}$ and $\rho_{u}$ denote the lengths; $\vartheta_{a}$ and $\vartheta_{u}$ denote the directions. Here, we define a control input space boundary function that gives the maximum length of action direction $\vartheta$ in the current car state, that is,
 \begin{equation}
 	\bar{\rho} = \Psi(v_x, \delta, \vartheta)
 \end{equation}

\begin{figure}[!ht]
\centering
  \includegraphics[width=0.60\linewidth]{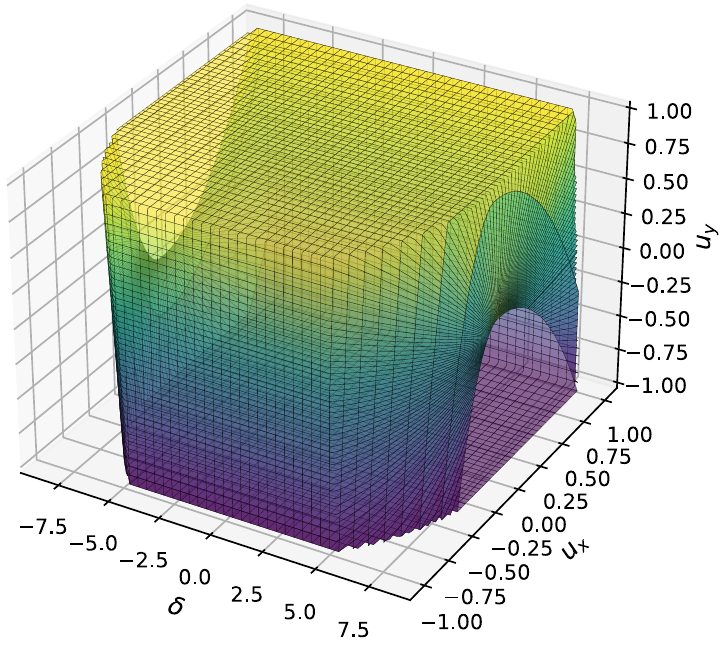}
  \caption{Admissible control input space at $v_x=15.4~\text{m/s}$ with full range of steering angle.}
  \label{fig_amspd3d}
\end{figure}
\begin{figure}[!ht]
\centering
  \includegraphics[width=0.65\linewidth]{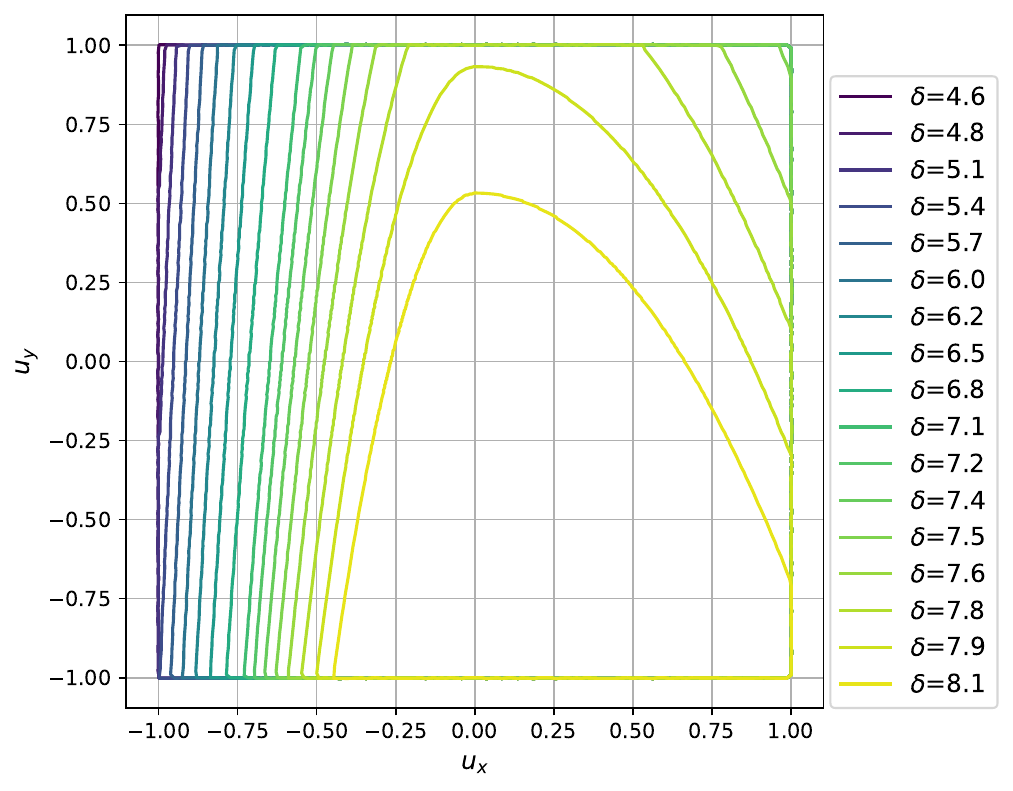}
  \caption{Boundaries of constrained control input space with different steering angles $\delta \in [4.6, 8.1]~\text{deg}$ at velocity $v_x=15.4~\text{m/s}$}
  \label{fig_amspd2d}
\end{figure}

In the following, we present a numerical method to determine the boundary function. Let $v_i = \{v_1, v_2, \ldots, v_{N_{v}}\}$ be $N_{v}$ evenly spaced velocity values over $[0, v_{\max}]$, and $v_{\max}$ is the maximum speed. Let $\delta_j = \{\delta_1, \delta_2, \ldots, \delta_{N_{\delta}}\}$ be $N_{\delta}$ evenly spaced front wheel steering angle values over $[-\delta_{\max}, \delta_{\max}]$, and $\delta_{\max}$ is the maximum steering angle. Let $\vartheta_k = \{\vartheta_1, \vartheta_2, \ldots, \vartheta_{N_{\vartheta}}\}$ be $N_{\vartheta}$ evenly spaced action vector directions over $(-\pi, \pi]$. Then, we iterate all combinations of the race car's state $(v_i, \delta_j)$ with all control input directions and lengths, and check the car's response to control inputs using the dynamic model. More specifically, for each car state $(v_i, \delta_j)$, we iterate all action directions. For a certain direction $\vartheta_k$, we apply the control input vector to the dynamic system while increasing the length $\rho$ till the control input fails to satisfy the tire friction constraint, and we could determine the maximum length $\bar{\rho}$ at car state $(v_i, \delta_j)$ and action direction $\vartheta_k$. Based on this sampling method, we obtain a look-up table of three dimensions to represent the boundary function, that is $\bar{\rho}_{i,j,k} = \Psi(v_i, \delta_j, \vartheta_k)$. To demonstrate the boundary function built from sampling, we visualize the boundary function of $v_x=15.4~\text{m/s}$ in Fig. \ref{fig_amspd3d}. Inside the 3D space is the admissible space for control inputs $[u_x, u_y]$ of different steering angle $\delta$. Fig. \ref{fig_amspd2d} gives another view of the control input boundaries at velocity $v_x=15.4~\text{m/s}$ with steering angle $\delta$ in range $[4.6, 8.1]$ deg. 

Due to the discrete nature of the boundary function, we cannot directly find the maximum length $\bar{\rho}$ for any $v_x$, $\delta$, and $\vartheta_a$, which are continuous values. Therefore, we use an off-the-shelf linear multidimensional interpolation method to approximate the maximum length. The approximated boundary function is denoted by $\hat{\rho} = \hat{\Psi}(v_x, \delta, \vartheta)$, where $\hat{\rho}$ is the approximate value of $\bar{\rho}$. Finally, we summarize the numerical action mapping procedure with the discrete boundary function in \textbf{Algorithm \ref{alg_AM}}. With the help of this numerical action mapping method, the action that could violate the constraint is mapped to a safe control input right inside the boundary. This method fundamentally prevents the race car from entering uncontrollable states while making full use of the maximum tire-road friction.



\begin{algorithm}[!ht]
\caption{ Numerical Action Mapping with Discrete Boundary Function}
\begin{algorithmic}[1] \label{alg_AM}
\STATE Load discrete boundary function $\bar{\rho}_{i,j,k} = \Psi(v_i, \delta_j, \vartheta_k)$
\STATE Input car speed $v_x$ and steering angle $\delta$
\STATE Input unconstraint virtual action $a_t=[a_x, a_y]$
\STATE Convert virtual action to polar coordinate form $[a_x, a_y]\rightarrow [\rho_a, \vartheta_a]$
\STATE Calculate maximum length $\hat{\rho} = \hat{\Psi}(v_x, \delta, \vartheta_a)$ using a multidimensional interpolation method on discrete boundary function
\IF {$\rho_a \leq \hat{\rho}$}
\STATE constraint control $u_t=[\rho_a, \vartheta_a]$
\ELSE 
\STATE constraint control $u_t=[\hat{\rho}, \vartheta_a]$
\ENDIF
\STATE Convert to Cartesian coordinate form $u_t = [\hat{\rho}, \vartheta_a] ~\text{or}~[\rho_a, \vartheta_a]\rightarrow [u_x, u_y]$ 
\STATE Output constraint control $u_t=[u_x, u_y]$
\end{algorithmic}
\end{algorithm}


%


\subsection{Implementation of RL Training with AM}

The AM mechanism can address the state-dependent constraint for a variety of policy gradient-based RL algorithms with parameterized policy and continuous action space, such as DDPG, TD3, PPO, SAC, etc. In this subsection, we incorporate the numerical approximation method of AM to the TD3 algorithm to train an autonomous race driving policy as an example.

\begin{figure}[!ht]
\centering
  \includegraphics[width=0.6\linewidth]{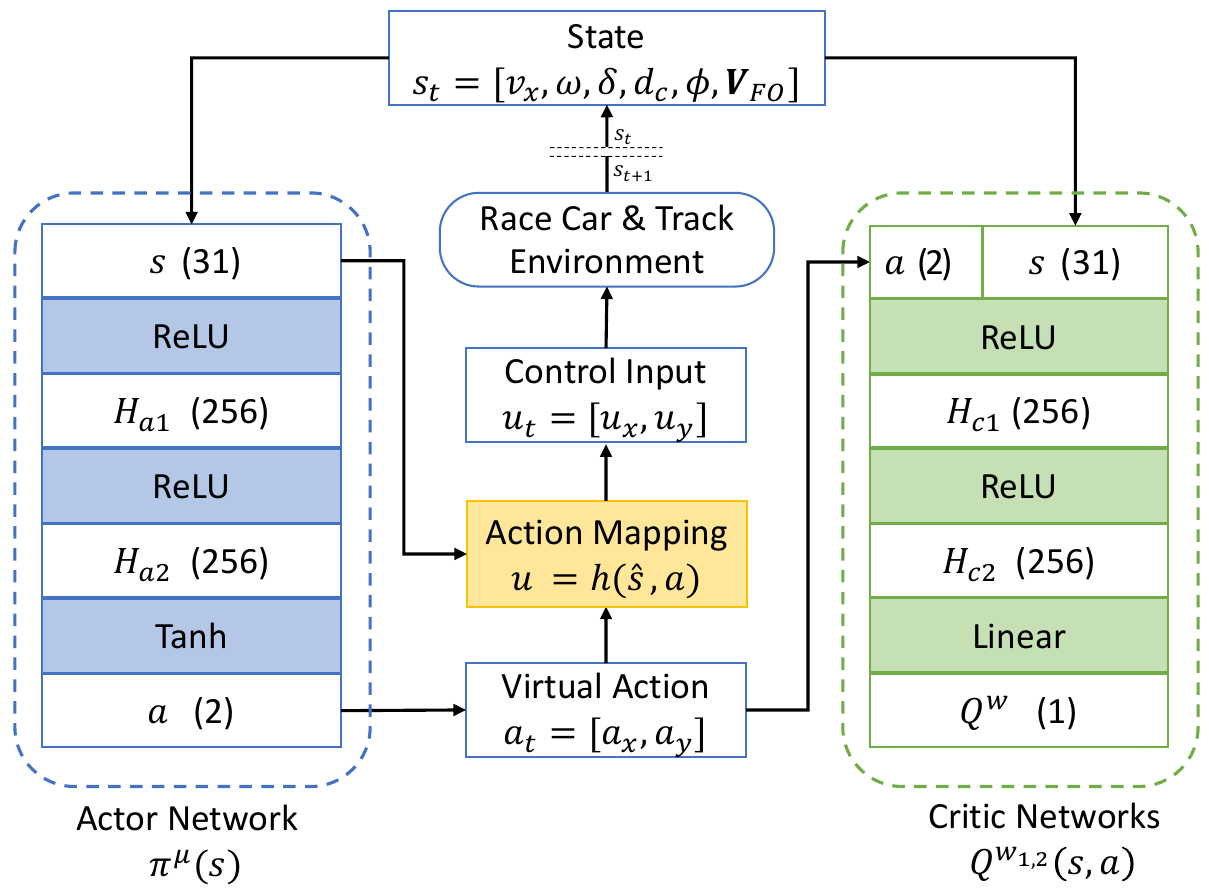}
  \caption{Diagram of TD3 with AM for race driving policy training and the structure of the actor and critic networks. }
  \label{fig_acnet}
\end{figure}

TD3 is a deterministic policy-based reinforcement learning algorithm that employs an actor-critic architecture. The actor and critic functions are represented by fully connected neural networks. A block diagram of the network structure is shown in Fig. \ref{fig_acnet}. The actor network, which represents the unconstrained virtual policy is denoted as $\pi^{\mu}(s)$, and $\mu$ is the network parameters. As illustrated in Fig. \ref{fig_acnet}, the actor network has two hidden layers, and each hidden layer contains $256$ hidden nodes with the ReLU activation function. For the output layer, we use the tanh function to limit the actions between $-1$ and $1$. The action network's target network shares the same structure as the action network, and it is denoted as $\pi^{\mu'}(s)$. The state-action value function $Q^{\pi^{\mu}}(s, a)$ is approximated by the critic networks. Different from the conventional DDPG algorithm, the TD3 algorithm introduces a pair of critic networks to mitigate the value function overestimation. The two critic  networks are denoted by $Q^{w_1}(s, a)$ and $Q^{w_2}(s, a)$. They share the same structure which is shown in Fig. \ref{fig_acnet}. The input of critic networks concatenates the state vector and the action vector. The hidden layers have the same structure as the actor network while the output layer uses a linear function. The target networks of the critic networks are denoted by $Q^{w'_1}(s, a)$ and $Q^{w'_2}(s, a)$. 

The critic networks are trained using a batch training method with a replay buffer. For each time step $t$, a state transition experience $e_t=(s_t, a_t, r_{t+1}, s_{t+1})$ is saved to the replay buffer $\mathcal{D}$. During each training iteration, a batch of experiences $(s_i, a_i, r_i, s'_i)_{i=1,2,\ldots,N}$ is randomly selected from the replay buffer. Here, $r_i$ and $s'_i$ are the reward received and state reached after taking $a_i$ at state $s_i$. Next, two critic networks are trained separately to minimize the TD-error using the loss function:
\begin{equation}
	\mathcal{L}(w_j) = \frac{1}{N}\sum^{N}_{i=1}\Big[y_i - Q^{w_j}(s_i, a_i)  \Big]^2,
\end{equation}
where $j=1,2$, and $y_i$ is the target value from the target critic networks, which is,
\begin{equation} \label{eq_target}
	y_i = r_i + \gamma \min_{j=1,2}Q^{w'_j}(s'_i,\pi^{\mu'}(s'_i)+ \epsilon).
\end{equation}
The target value for the critic networks is from the smaller value between two target networks, which reduces the overestimation of the state-action value. $\epsilon$ is a small noise sampled from a clipped Gaussian distribution $\text{clip}(\mathcal{N}(0,\sigma_{\epsilon}^2),-c,c)$, which is used to avoid overfitting. After the target values are determined, the gradients is given as:
\begin{equation} \label{eq_critic1}
	\nabla_{w_j}\mathcal{L}(w_j) = \frac{1}{N}\sum^{N}_{i=1}\Big[y_i-Q^{w_j}(s_i,a_i)\Big]\nabla_{w_j}Q^{w_j}(s_i,a_i).
\end{equation}
The parameters of two critic networks are adjusted according to:
\begin{equation} \label{eq_critic2}
	w_j \leftarrow w_j - \alpha_w \nabla_{w_j} \mathcal{L}(w_j),
\end{equation} 
where $\alpha_w$ is a small updating rate. 

The policy's performance is assessed based on the expected return starting from the initial time step, that is, $J(\mu)=\mathbb{E}_{\pi^{\mu}}[R_1]$. According to the deterministic policy gradient theorem \cite{silver2014deterministic}, the gradient of the policy performance is given by:
\begin{equation}
	\nabla_{\mu}J(\mu) = \mathbb{E}_{\pi^{\mu}}\Big[\nabla_{\mu}\pi^{\mu}(s)\nabla_{a}Q^{\pi^{\mu}}(s,a)|_{a=\pi^{\mu}(s)}\Big].
\end{equation}
Then, the actor network is trained using the approximated gradient of the performance function based on the critic network $Q^{w_1}(s,a)$ and the same batch of transition experiences. The approximation of the policy gradient is given by:
\begin{equation}\label{eq_actor1}
	\nabla_{\mu}J(\mu) \approx \frac{1}{N}\sum^{N}_{i=1}\nabla_{\mu}\pi^{\mu}(s_i)\nabla_{a}Q^{w_1}(s_i,a_i)|_{a_i=\pi^{\mu}(s_i)}.
\end{equation}
The virtual control policy is improved by updating the parameter $\mu$ using gradient ascent with a small updating rate $a_\mu$:
\begin{equation} \label{eq_actor2}
	\mu \leftarrow \mu + \alpha_\mu \nabla_\mu J(\mu)
\end{equation}
After each training iteration, the target networks are soft-updated with small update rates $\beta_w$ and $\beta_{\mu}$:
\begin{align}
	w'_j &\leftarrow \beta_w w_j + (1-\beta_w)w'_j \label{eq_critictarget}\\
	\mu' &\leftarrow \beta_{\mu} \mu + (1-\beta_{\mu})\mu' \label{eq_actortarget}
\end{align}

The overall training procedures of the race driving policy with TD3-AM are summarized in \textbf{Algorithm \ref{alg_TD3AM}}. In line 9, a Gaussian noise $n_t \sim \mathcal{N}(0, \sigma^2_{\mu})$ is added to the virtual action to promote exploration. In lines 14-20, a delayed policy update method is utilized to further improve the stability of the training process. In particular, updates to the actor network and the target networks occur after a set number ($T_\text{delay}$) of updates to the critic networks. Furthermore, the policy is saved and evaluated at a certain interval of training iterations. 

\begin{algorithm}[!ht]
\caption{Race Driving Policy Training with TD3-AM}
\begin{algorithmic}[1] \label{alg_TD3AM}
\STATE Randomly initialize the parameters the actor network, twin critic networks, and their target networks.\\
\STATE Initialize replay buffer $\mathcal{D}$\\
\STATE Load race driving simulation environment\\
\FOR{episode = $1$ \TO MaxEpisode}
	\STATE Set initial car state\\
	\STATE Observe initial state $s_1$\\
	\FOR{time step $t = 1$ \TO MaxStep}
		\STATE Generate virtual action $a_t = \pi^{\mu}(s_t) + n_t$\\
		\STATE Mapping to real control input $u_t=h(\hat{s}_t, a_t)$\\
		\STATE Apply control input $u_t$ to car dynamic model\\
		\STATE Observe new state $s_{t+1}$ and receive reward $r_{t+1}$\\
		\STATE Store the transition $(s_t, a_t, r_{t+1}, s_{t+1})$ in $\mathcal{D}$
		\STATE Select a batch of $N$ experiences randomly from $\mathcal{D}$
		\STATE Calculate target values according to (\ref{eq_target})
		\STATE Update critic networks according to (\ref{eq_critic1}) and (\ref{eq_critic2})
		\IF {$t$ mod $T_{\text{delay}}=0$}
			\STATE Update actor network according to (\ref{eq_actor1}) and (\ref{eq_actor2})
			\STATE Update critic target networks according to (\ref{eq_critictarget})
			\STATE Update actor target network according to (\ref{eq_actortarget})
		\ENDIF
		\IF {car drives off-track or wrong-way} 
			\STATE \textbf{break}
		\ENDIF
	\ENDFOR	
\ENDFOR
\end{algorithmic}
\end{algorithm}

\section{Simulations and Results} \label{sec_simulation}
In this section, the proposed RL-based race driving strategies are evaluated in our built race simulation environment. In the following, we first introduce the simulation environment and training details. Then, the evaluation results are presented and discussed. 

\subsection{Simulation Environment and Training Details}
The simulation environment is developed following the vehicle model given in Section \ref{sec_model}. The Runge-Kutta four-order (RK4) method is used to numerically solve the differential equations. The simulation time step is $0.01$s. The car model used in the simulation is an all-electric mid-size sedan. The physical parameters of the car model are given in Table \ref{tb_car}. 


\begin{table}[!ht]
\centering
\caption{ Car Model Parameters}
\begin{tabular}{lrr}
\toprule
Parameter & Value & Unit\\
\midrule
Vehicle mass $m$ & $1860$ & kg \\
Front axle distance $l_f$ & $1.17$ & m \\
Rear axle distance $l_r$ & $1.77$ & m \\
Tire rolling radius $R_w$ & $0.31$ & m \\
Tire corner stiffness $C_{\alpha_f}, C_{\alpha_r}$ & $54,500$ & N/rad \\
Tire rolling resistance coefficient $f_r$ & $0.015$ & - \\
Maximum steering angle & $35$ & deg \\
Yaw moment of inertia $I_{z}$ & $4000$ & kg$\cdot \text{m}^2$ \\
Aerodynamic drag coefficient $C_d$ & $0.3$ & - \\
Density of air $\rho_A$ & $1.2258$ & kg/$\text{m}^3$ \\
Vehicle frontal area $A_f$ & 2.05 &  $\text{m}^2$\\
Maximum motor power $P_{\text{max}}$ & $125$ & kW \\
Motor torque coefficient $K_m$ & $1,550$ & N$\cdot\text{m}$ \\
Braking force coefficient $K_b$ &  $16,422$ & N \\
Maximum friction coefficient $\mu_{\max}$ & $1.15$ & - \\
Acceleration due to gravity $g$ & $9.81$ & $\text{m}/\text{s}^2$ \\
\bottomrule
\label{tb_car}
\end{tabular}
\end{table}

\begin{figure}[ht]
\centering
  \subfloat[Track-A (length: 860m; width: 20m)]{\includegraphics[width=0.35\linewidth]{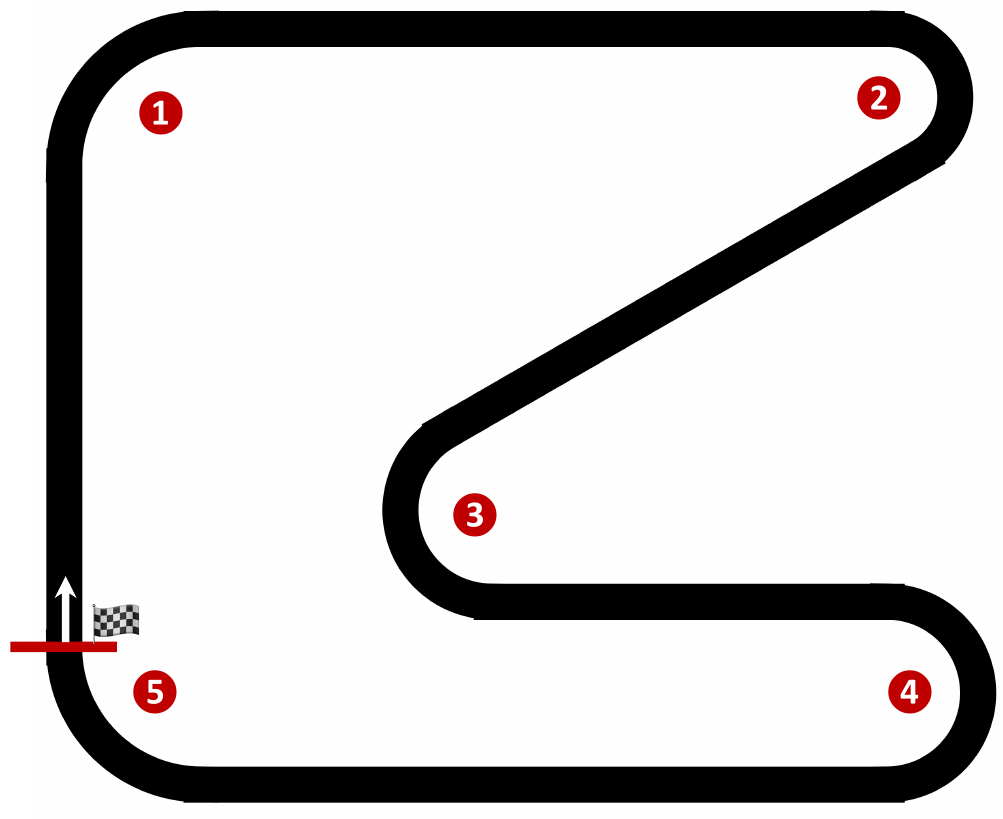}}
  \hspace{0.5cm}
  \subfloat[Track-B (length: 1400m; width: 10m)]{\includegraphics[width=0.6\linewidth]{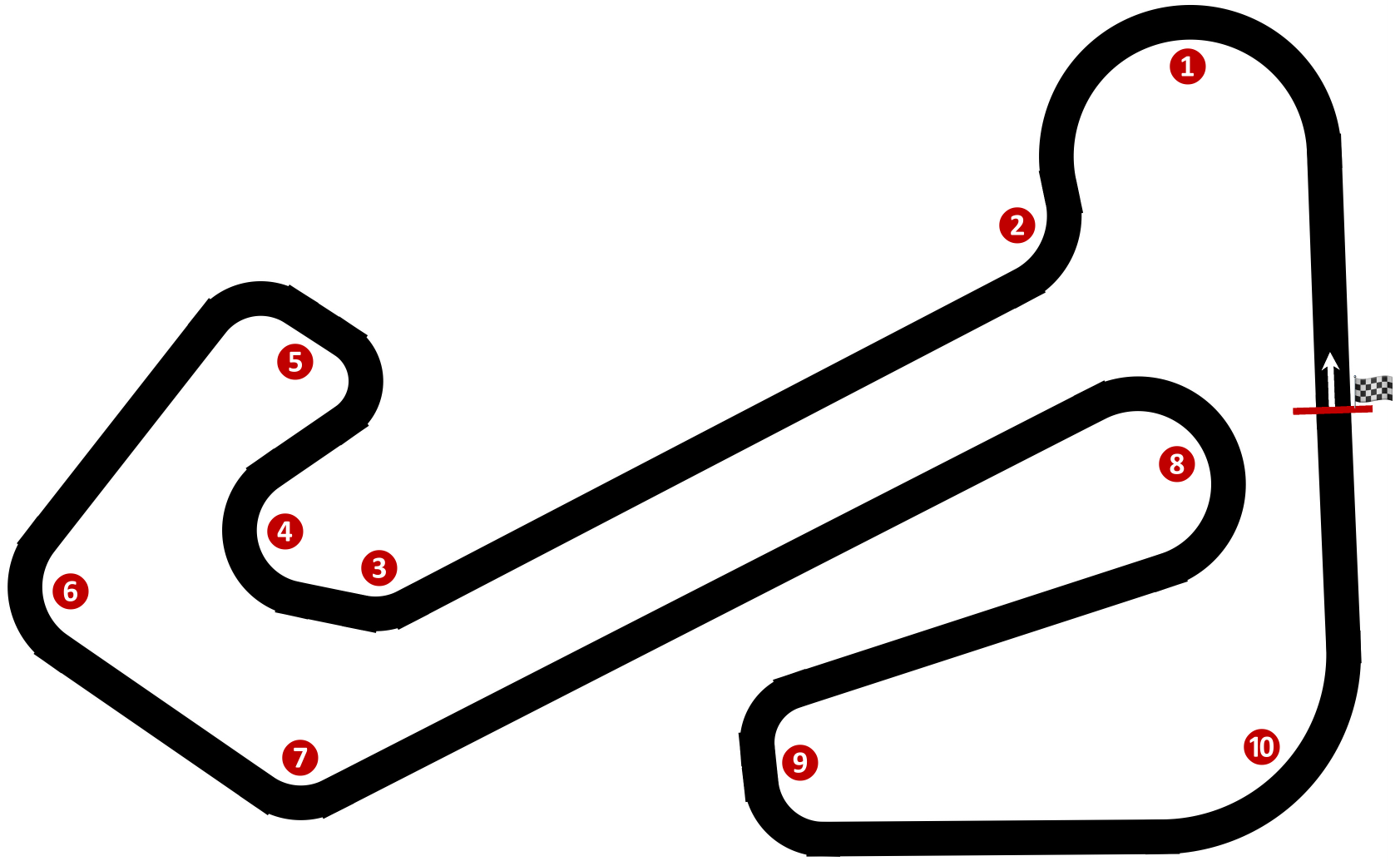}}
  \caption{Layout of two race tracks in the racing simulation environment}
  \label{fig_tracks}
\end{figure}

We build two race tracks for race driving policy training and evaluation. The layout of track-A is given in Fig. \ref{fig_tracks}(a). It is a simple testing track with only five corners. The length and width of track-A are $860$m and $20$m respectively. The layout of track-B is given in Fig. \ref{fig_tracks}(b). The layout is modeled after the Ruisi Circuit located in Beijing, China. Track-B has ten corners. The total length is $1400$m and the width is $10$m. In Fig. \ref{fig_tracks}, the finish line and running direction of the tracks are marked with a red line across the track and a white arrow respectively.

\begin{figure}[!ht]
\centering
  \includegraphics[width=0.46\linewidth]{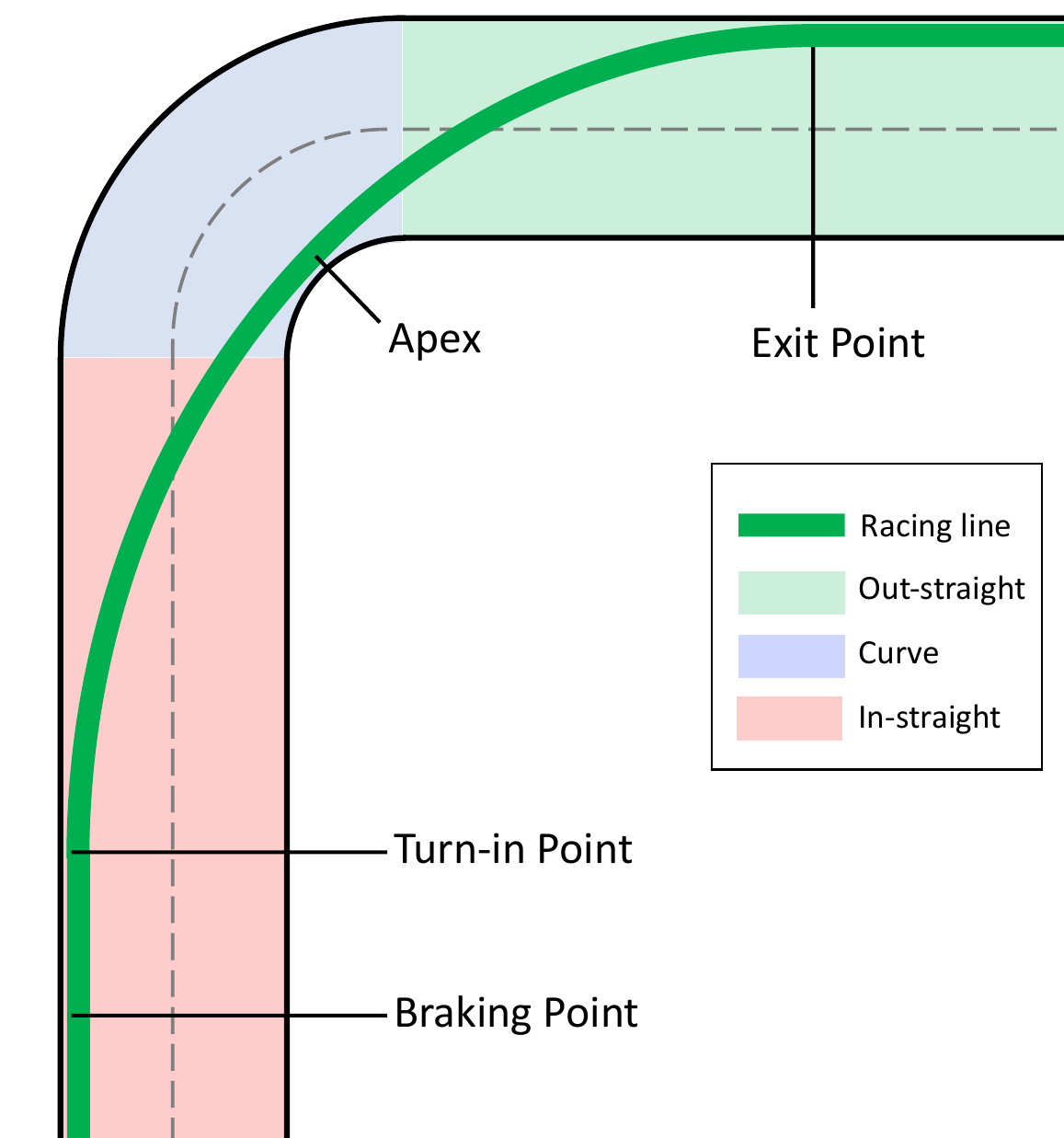}
  \caption{A right angle turn corner with racing line and key points.}
  \label{fig_corner}
\end{figure}

A typical race track can be regarded as a sequence of basic corners connected together. Therefore, the driving skill of cornering is the key to minimizing the lap time. As shown in Fig. \ref{fig_corner}, a basic corner can be divided into three parts: in-straight (red part), curve (blue part), and out-straight (green part). The solid green line indicates the racing line, which is the theoretical fastest line through the corner. Obviously, the racing line significantly reduces the tightness of the corner by using the in-straight and the out-straight parts, which allows for the highest speed possible to run through this corner. There are four key points on the racing line. The braking point is the position to start applying the brake before the corner. The turn-in point is the position to start steering into the corner. The apex is located on the inside of a corner, and it is the aiming point after the turn-in point. The apex is also the position to start acceleration. The exit point is where the car once again reaches the outside of the corner, and it is also the aiming point after the car passes the apex. The ideal racing line and key points depend on the curvature of the corner, the condition of the previous corner and the following corner, and the handling performance of the car being driven. 

We conduct the following simulation experiments on a workstation with Ubuntu 20.04 operating system, Intel Core i9-13900k CPU, 32GB RAM, and NVIDIA  GeForce RTX 4090 GPU. The neural networks for RL algorithms are built with the PyTorch framework and implemented on the GPU.

The driving policy is trained following \textbf{Algorithm \ref{alg_TD3AM}}. To implement the numerical AM mechanism, we first determine the discrete boundary function based on the race car's dynamics with friction constraint. The numbers of the discretization points are set as: $N_v=N_\delta=N_\vartheta=200$. The discretization step sizes for speed, steering angle, and action vector angle are $0.15$ m/s, $0.35$ deg, and $1.8$ deg, respectively. Then, we could use the AM mechanism in policy training following \textbf{Algorithm \ref{alg_AM}}. The training parameters are listed in Table \ref{tb_rlparam}. At the beginning of each training episode, we place the car on a random position in the straight part of the track with a random speed $v_x \in [0, 30]$ . The initial $\omega$, $\phi$, and $\delta$ are all zero. For each time step, the states are normalized to $[0,1]$ or $[-1,1]$ with respect to their minimum and maximum values. For the forward-observation feature state $\textbf{V}_{\text{FO}}$, we utilize $12$ forward observation points at varying distances: $d_i = \{10, 20, 30, 40, 60, 80, 100, 120, 140, 160, 180, 200 \}$ meters. The maximum duration for one episode is $10,000$ steps, equivalent to $100$ seconds.

\begin{table}
\centering
\caption{Training Parameters}
\begin{tabular}{p{0.6\linewidth} >{\raggedleft\arraybackslash}p{0.2\linewidth}}
\toprule
Parameter & Value \\
\midrule
Discount factor $\gamma$ & $0.99$ \\
Initial learning rate for critic network $\alpha_{w}$ & $0.0003$ \\
Initial learning rate for actor network  $\alpha_{\mu}$ & $0.0003$ \\
Updating rate for target critic network $\beta_{w}$ & $0.005$ \\
Updating rate for target actor network $\beta_{\mu}$ & $0.005$ \\
Batch size $N$ & $256$ \\
Replay buffer size $M$ & $1,000,000$ \\
Exploration noise variance $\sigma_\mu$ & $0.1$ \\
Policy smoothing variance $\sigma_\epsilon$ & $0.2$ \\
Update delay step $T_{\text{delay}}$ & $2$ \\
\bottomrule
\label{tb_rlparam}
\end{tabular}
\end{table}

To further explore the capabilities of the proposed TD3-AM algorithm, we introduce several comparative approaches, including the conventional TD3 algorithm, the proximal policy optimization (PPO) algorithm, and the safety layer technique (SL). Given that TD3 operates as an off-policy RL algorithm, we opt for the on-policy PPO algorithm as an additional baseline RL algorithm, and we also apply AM to the PPO algorithm. In our experiments, the PPO algorithm is implemented with the Stable-Baselines3 library \cite{stable-baselines3}. The structure of the actor and critic networks and the batch size used for PPO are the same as the TD3-AM. Other training parameters are also carefully tuned. The safety layer is a state-of-the-art safe exploration technique for RL with state constraints  \cite{dalal2018safe}. As we introduced before, SL shares a similar idea with AM, which is adjusting the original action from the actor network to satisfy the constraints. Therefore, it is also compatible with both on-policy and off-policy RL algorithms. In the following experiments, SL is built following the procedures given in \cite{dalal2018safe}. We first collect transition data from the simulation environment where the race car is randomly initialized and given random actions. The transitions of constraint violations are specially marked.  Then, the immediate-constraint function of SL is approximated with a two-hidden-layer neural network based on the collected data through supervised learning. After that, we apply SL to both TD3 and PPO algorithms.

\subsection{Evaluation Results on Track-A}

 For the evaluation on track-A, we train and test all six approaches mentioned above, which are: 1) TD3; 2) TD3-SL; 3) TD3-AM; 4) PPO; 5) PPO-SL; 6) PPO-AM. Each approach is trained for $20$M iterations.   After every $10$k iterations, the driving policy is evaluated by an evaluation episode where the race car starts from the finish line and runs for $100$s. The accumulated reward and lap time of the episode is recorded. For each approach, we perform $10$ independent training trials starting with randomly initialized actor and critic networks.
 
\begin{figure}[ht]
\centering
  \includegraphics[width=\linewidth]{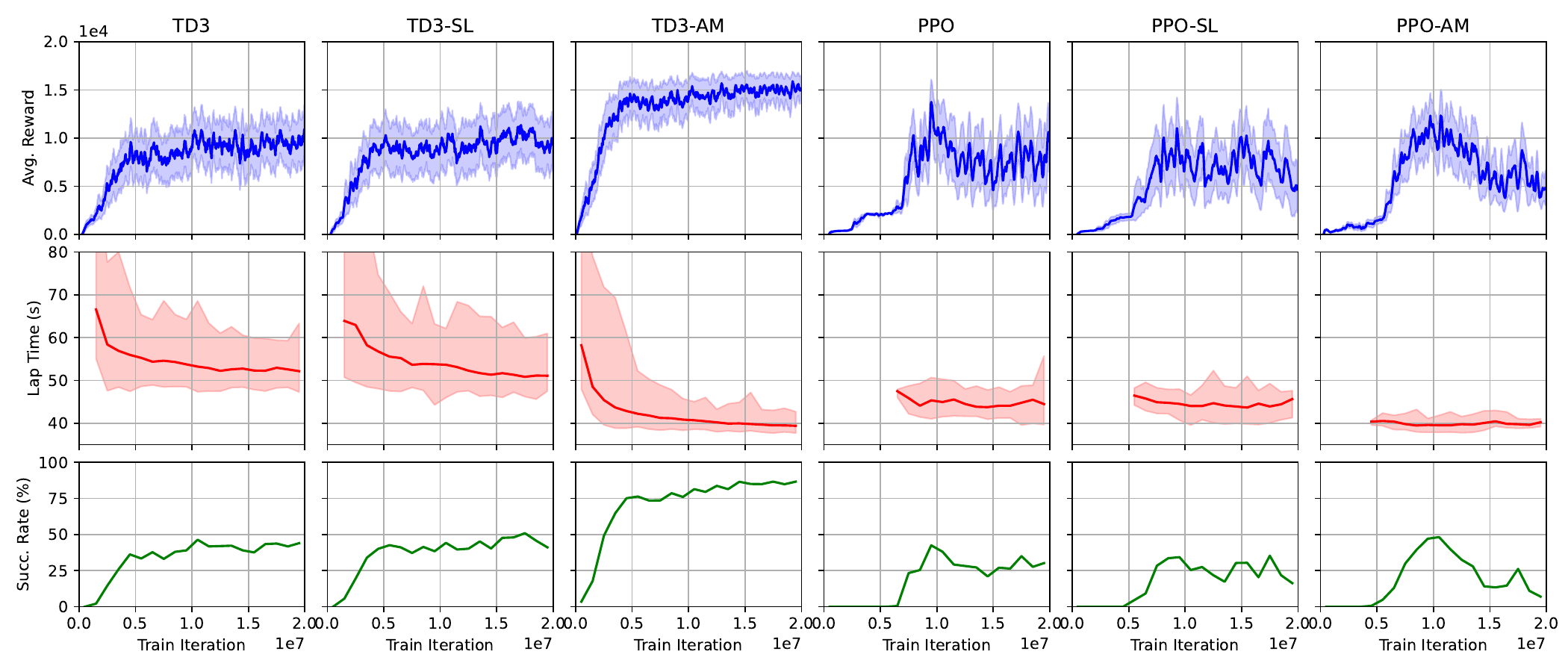}
  \caption{Evaluation result of six approaches from $10$ independent trials. First row: average rewards with standard deviations; Second row: max, median, and min lap time; Third row: success rate of finish two laps.}
  \label{fig_reward}
\end{figure}

Figure \ref{fig_reward} (first row) displays the learning progress, as indicated by the average accumulated rewards of the evaluation episodes. The solid lines and shaded areas depict the average values and standard deviations of the rewards for the identical evaluation episode across 10 trials. The curves show that the learning progress of the TD3-based approaches is more stable and consistent, while the PPO-based approaches are not very stable. Moreover, the TD3-AM approach achieves the highest average reward.
  
 In motorsport, the lap time of a `flying lap' is a major criterion to evaluate a race car's performance and the skill of a race driver. The `flying lap' starts when the car crosses the finish line of the previous lap at a high rate of speed. In our evaluation episode, if the driving policy could finish at least two laps without any fault, the second lap can be regarded as a `flying lap' and this episode is identified as a `success episode'. To demonstrate the performance improvement of the driving policies during the learning process, we aggregate the results of every $10$ evaluation episode from $10$ trials as a group. The lap time and success rate of all groups are given in the second and third rows of Fig. \ref{fig_reward}. The median values and the max/min value of lap time in each evaluation group are shown by solid lines and shaded areas respectively. If there is no success episode in one evaluation group, the lap time is not shown. The success rate of the TD3-AM policy finally reaches $90\%$, while other policies are all below $50\%$. The PPO-based policies obtain lower success rates than the TD3-based policies, and the PPO-based policies also take more training iterations before having a success episode. However, for the lap time, the PPO and PPO-SL policies obtain a shorter time than their corresponding TD3 and TD3-SL policies. More importantly, the driving policies with AM achieve further shorter lap time than other policies.
 
Furthermore, we introduce other criteria for comparison, including 1) the best lap time achieved in all evaluation episodes; 2) the average cumulative number of episodes terminated due to the friction constraint violation; and 3) training speed in seconds per 1k iteration. These values are listed in Table \ref{tb_tracka}. From the best lap time, we find that TD3 and PPO approaches make significant improvements after using AM. The best lap times of TD3-AM and PPO-AM are $22\%$ and $5\%$ shorter than their corresponding baselines. In comparison, the improvements by introducing SL are relatively smaller. The TD3-AM driving policy achieves the best lap time of all policies. From the number of constraint violations, we observe that introducing SL could reduce the number of friction constraint violations by half, and the AM mechanism successfully avoids any violations. The training speed of the PPO-based approaches is approximately three times faster than the TD3-based approaches. Applying AM and SL to TD3 and PPO prolongs the training time about $0.4$s and $0.2$s slower per $1$k iteration. Note that processing the action mapping or safety layer of a single step takes less than $0.2$ms.
 
Although both SL and AM are designed to address the constraint, only the AM mechanism achieves zero violation in all episodes. The main reason is that the numerical AM mechanism is more efficient at handling constraints with complex input coupling and nonlinear dynamics. From the vehicle model given above, the friction constraint function includes nonlinear couplings among speed, steering angle, and control inputs. To deal with the constraint, the SL technique approximates the friction constraint function with a linear model with respect to $g(s_c)^{T}[a_x, a_y]$, where the coefficient $g(s_c)$ is a function of car motion states $s_c$, extracted with a pre-trained neural network. Then, this constraint function is solved via an analytical optimization method. This method works well for decoupled systems. However, when there is coupling among inputs, this form of linear approximation does not fit the actual dynamics. We observe that the fitting accuracy of network $g(s_c)$ is relatively low from the pre-training, and that makes some of the actions fail to satisfy the constraint. Nevertheless, SL still works in some areas of the state space and effectively reduces the number of constraint violations.

\begin{table}[ht]
\centering

\caption{ Best lap time, constraint violation, training speed comparison of six approaches in track-A}
\begin{tabular}{lcccc}
\toprule
Driving Policy & \makecell[c]{Best Lap Time \\ (s)}  & \makecell[c]{Number of \\Constraint Violation} & \makecell[c]{Training Speed \\ (s/1k iteration)} \\
\midrule
TD3    & 47.28      & 252     & 1.186\\
TD3-SL & 44.26      & 123     & 1.572\\
\rowcolor[gray]{.9} TD3-AM & 36.94      & 0       & 1.579 \\
PPO    & 39.67      & 304     & 0.441 \\
PPO-SL & 39.54      & 135     & 0.644 \\
PPO-AM & 37.76      & 0       & 0.620 \\ 
\bottomrule
\label{tb_tracka}
\end{tabular}
\end{table}
 
\begin{figure}[!ht]
\centering
  \includegraphics[width=0.9\linewidth]{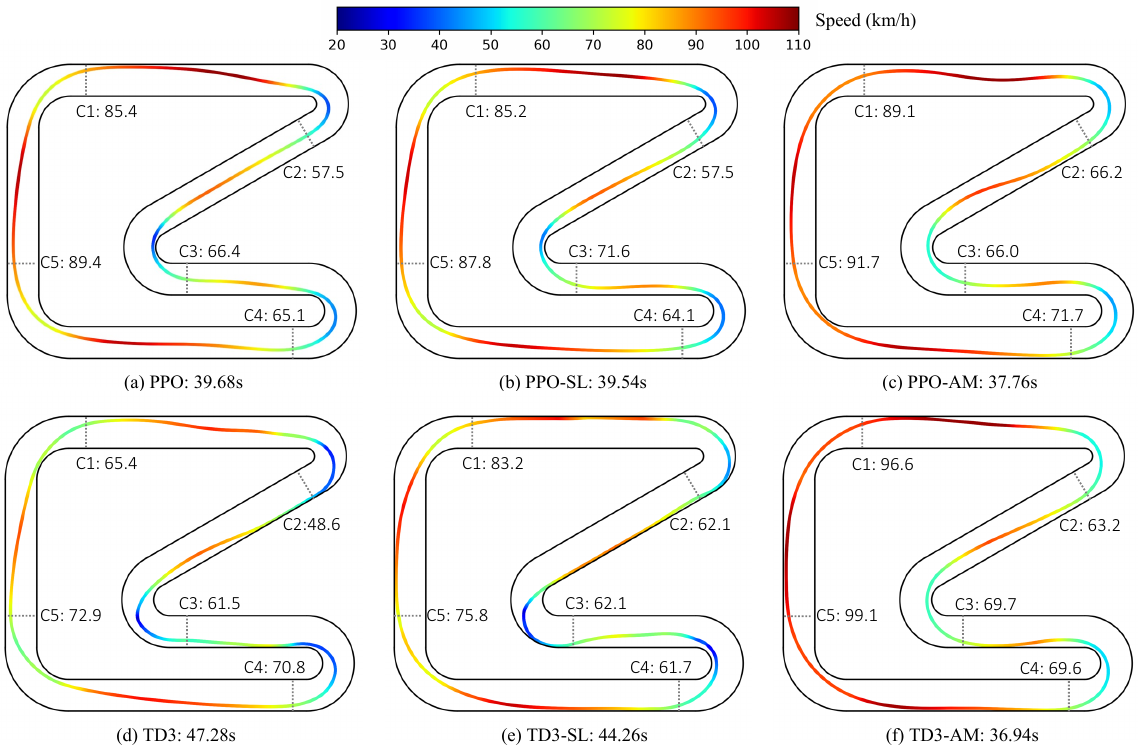}
  \caption{Best flying lap trajectories of six race driving policies in track-A. The speed is illustrated by the color of the trajectory.}
  \label{fig_fulltracktraj}
\end{figure}

The trajectories of the fastest flying lap by all six driving policies are shown in Fig. \ref{fig_fulltracktraj}. The speed is illustrated by the color of the trajectory. The speed at the exit point of corners C1-C5 is marked on the figure. From the trajectories and speed, we find that all driving policies have mastered some basic skills of race driving. The policies have learned to decelerate upon approaching a corner and accelerate after exiting a corner. The policies also have learned to use the racing line to drive at a higher speed through a corner. Comparing these trajectories reveals that the PPO-AM and TD3-AM policies exhibit smoother and more extended trajectories than others. The speeds at the exit points are consistently higher than those achieved by other policies. Both PPO-AM and TD3-AM policies showcase excellent driving skills, with highly similar trajectories. 


\begin{figure}[!ht]
\centering
  \includegraphics[width=0.75\linewidth]{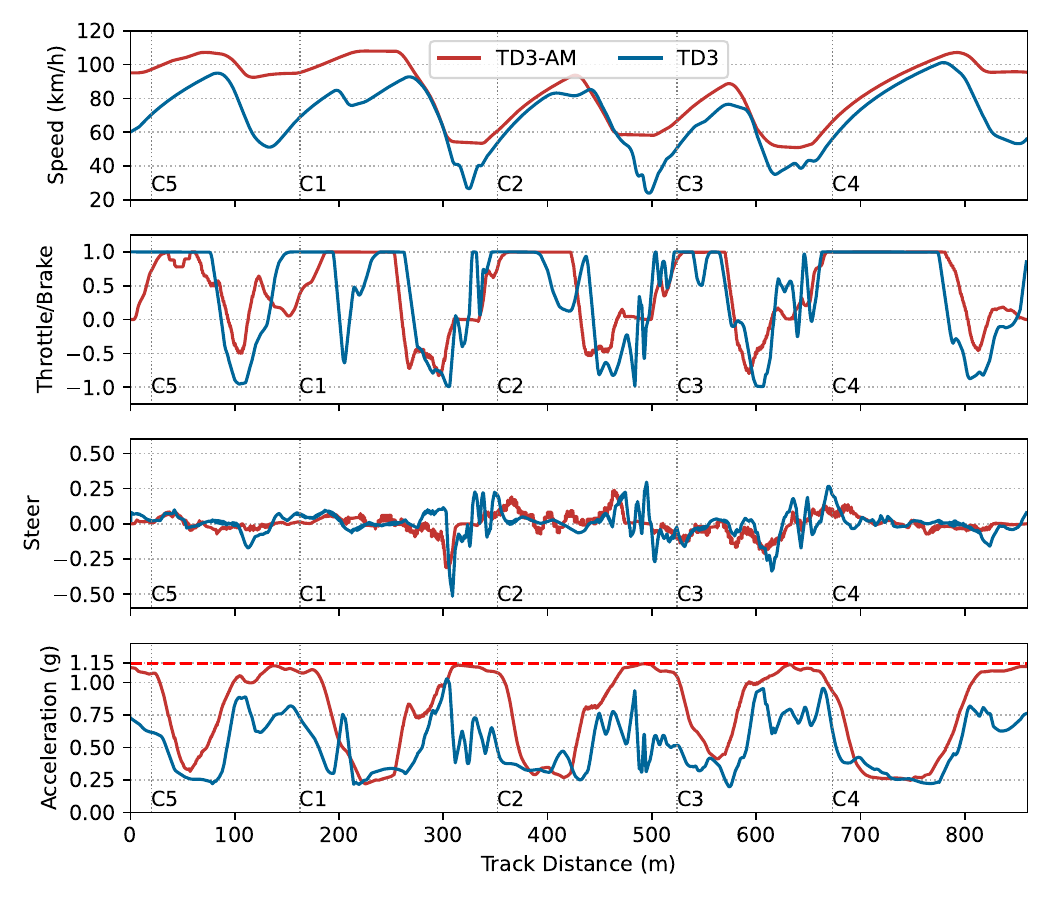}
  \caption{Speed, throttle/brake control, steering control, and resultant acceleration of TD3-AM and TD3 driving policies in track-A.}
  \label{fig_fulltrackstate}
\end{figure}

To comprehensively demonstrate the driving behavior of the learned policies, we show the speed, throttle/brake control, steering control, and resultant acceleration in Fig. \ref{fig_fulltrackstate}. Note that the $x$-axis denotes the track distance, which is the distance from the finish line to the vehicle's position along the centerline of the race track. The exit points of corners C1-C5 are marked in Fig. \ref{fig_fulltrackstate}. From the speed and throttle/brake curves, the TD3-AM policy tends to maintain a proper speed in corners. Differently, the TD3 policy continuously applies the brake, which makes the car run unnecessarily slow in corners. Furthermore, in corners C2, C3, and C4, the TD3 policy outputs fast-changing throttle/brake and steering control signals, which is harmful to the car's balance in real driving. In comparison, the control signals from the TD3-AM policy are much smoother. From the resultant acceleration, both driving policies satisfy the constraint of friction, which is given by a red horizontal line ($1.15g$) in the figure. In each corner, the acceleration of the TD3-AM policy is closer to the limit. That means the TD3-AM policy could better maximize the maneuverability of the vehicle. 

For conventional RL-based approaches (TD3 and PPO), constraint violations are penalized in the reward function, discouraging control policies from approaching the friction limit. The reward-shaping solution non-equivalently transforms the original optimization problem with constraints into a multi-objective optimization problem, where the constraint conditions become part of the penalty terms, which makes the control policy conservative. In comparison, introducing the AM mechanism directly prevents exceeding the friction limit from the perspective of action space without changing the objective of the optimization problem. Our method reconciles the contradiction between the need for higher speed and satisfying the friction constraint in turning maneuvers. In summary, the AM mechanism significantly improves the efficiency of RL algorithms in optimizing the race driving policy. 


\subsection{Evaluation Results on Track-B}

 The capability of the proposed TD3-AM driving policy is further evaluated on track-B. The generalization ability of the AM mechanism is also demonstrated and discussed. The corners in track-B are more complex and volatile compared to track-A, with the track width being only half of track-A. The six approaches are trained and evaluated on Track-B with identical training configurations, with the distinction that each approach is trained for 50M iterations instead of 20M. The learned driving policies are also evaluated every 1k iteration. However, both the TD3 and TD3-SL policies fail to achieve a satisfactory driving policy capable of completing a lap. The best lap times for the other four policies are listed in Table \ref{tb_laptimeb}. Similar to the results on track-A, the introduction of the AM mechanism also significantly enhances performance on track-B. Given the similarities in driving behaviors between PPO-AM and TD3-AM, we focus on demonstrating the TD3-AM driving policy in the following. 

\begin{table}[!ht]
\centering
\caption{ Best lap time comparison in Track-B}
\begin{tabular}{lc}
\toprule
Driving Policy & Best Lap time (min:sec) \\
\midrule
PPO  & 1:26.23 \\
PPO-SL & 1:18.38 \\
PPO-AM & 1:09.55 \\
 \rowcolor[gray]{.9} TD3-AM & 1:04.16 \\
\bottomrule
\label{tb_laptimeb}
\end{tabular}
\end{table}



\begin{figure}[!ht]
\centering
  \includegraphics[width=0.7\linewidth]{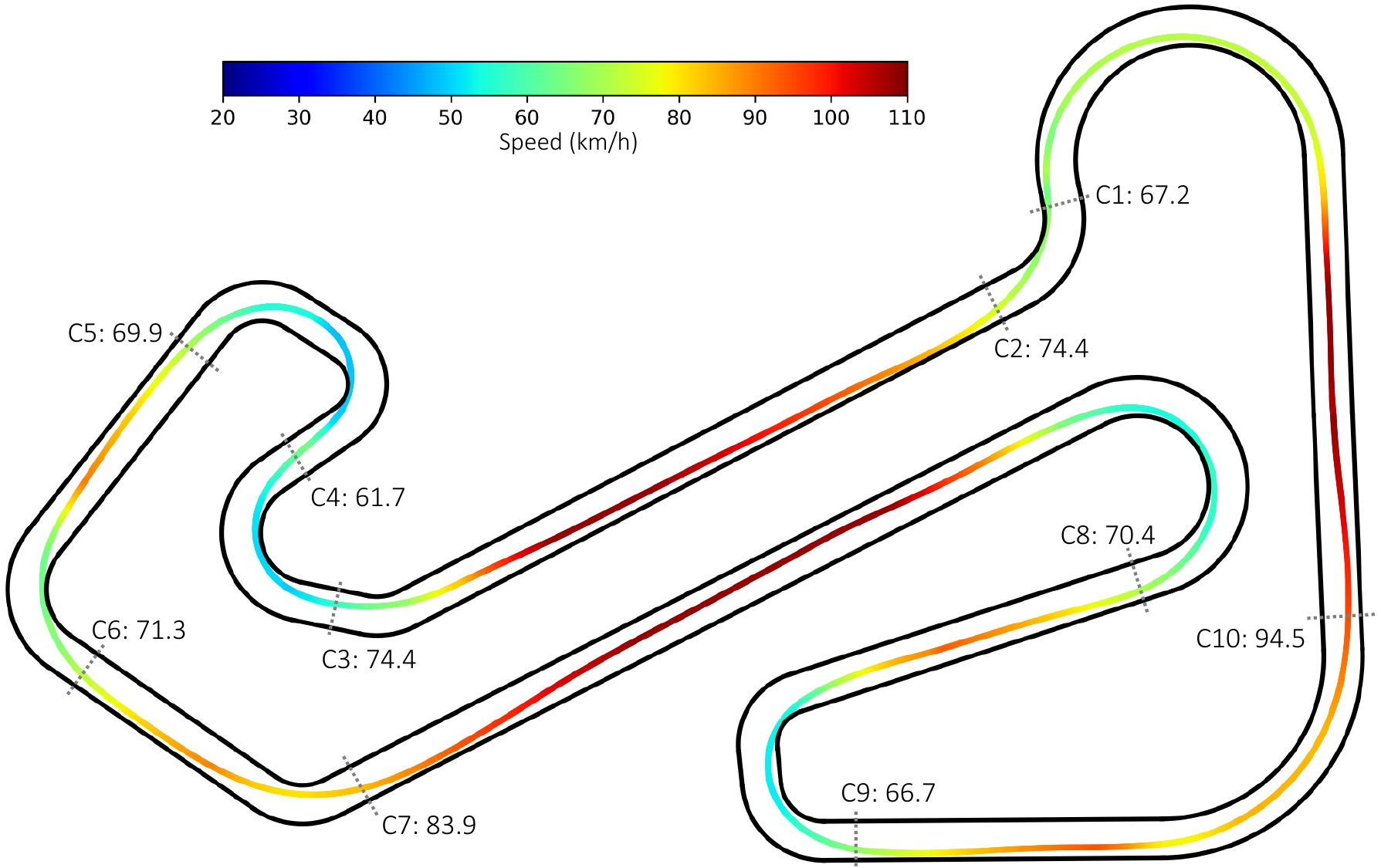}
  \caption{Flying lap trajectory by TD3-AM driving policy in track-B. The speed is illustrated by the color of the trajectory.}
  \label{fig_newtracktraj}
\end{figure}

The trajectory of the fastest flying lap by the trained TD3-AM driving policy is shown in Fig. \ref{fig_newtracktraj}. The color of the trajectory illustrates the speed. The speeds at the exit points of corners C1-C10 are marked in the figure. Although track-B is much more difficult than track-A, the TD3-AM algorithm still has successfully mastered the driving skills for track-B. In consecutive corners C3-C7 and hairpin corners C8 and C9, the driving agent utilizes the full width of the track to minimize corner curvature, resulting in a remarkably smooth trajectory. The overall trajectory closely aligns with the theoretical optimal race line.

\begin{figure}[!ht]
\centering
  \includegraphics[width=0.6\linewidth]{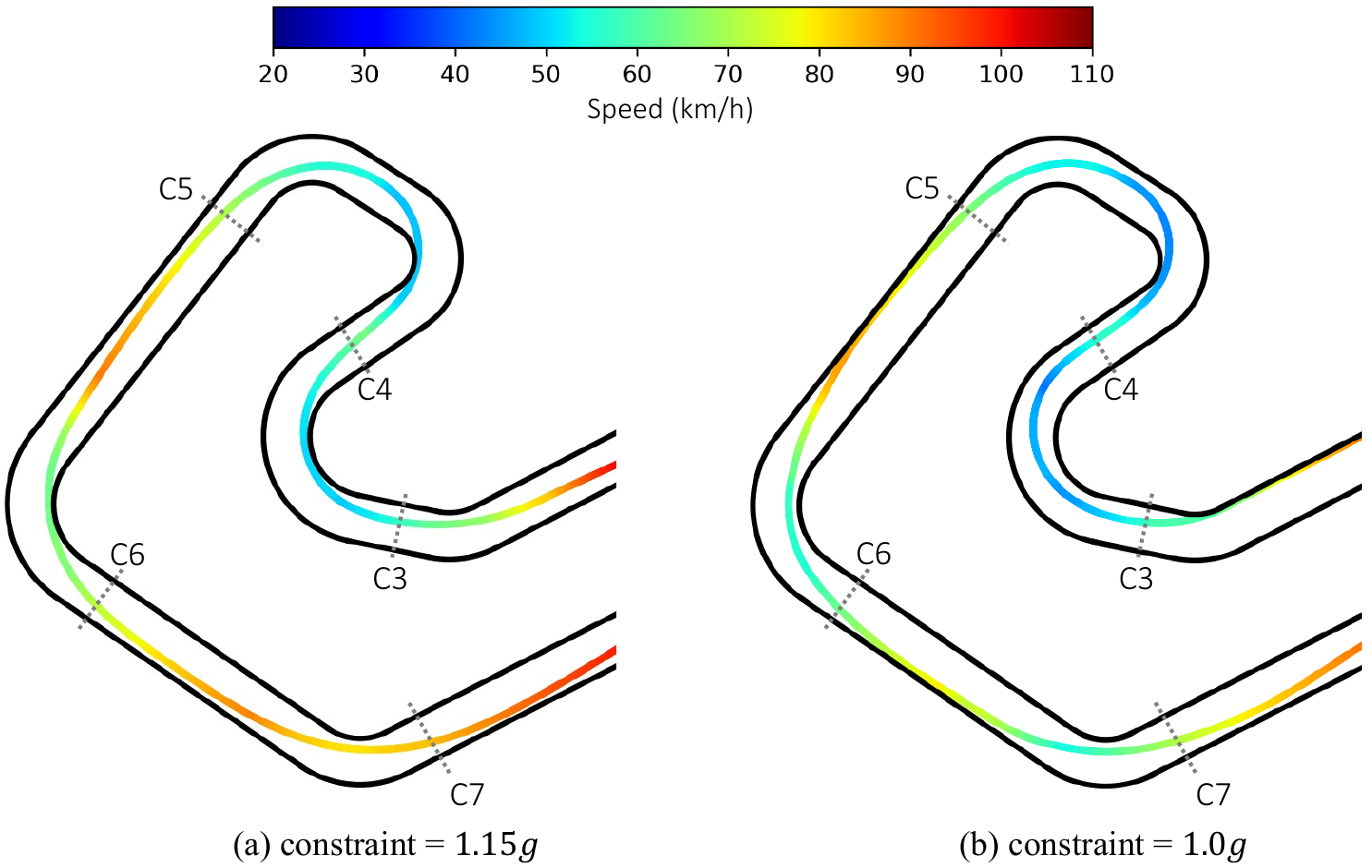}
  \caption{Flying lap trajectory comparison by TD3-AM driving policy with different friction constraints ($1.15g$ and $1.0g$) in track-B corners C3 to C7.}
  \label{fig_newtracktraj2}
\end{figure}

\begin{figure}[!ht]
\centering
  \includegraphics[width=0.75\linewidth]{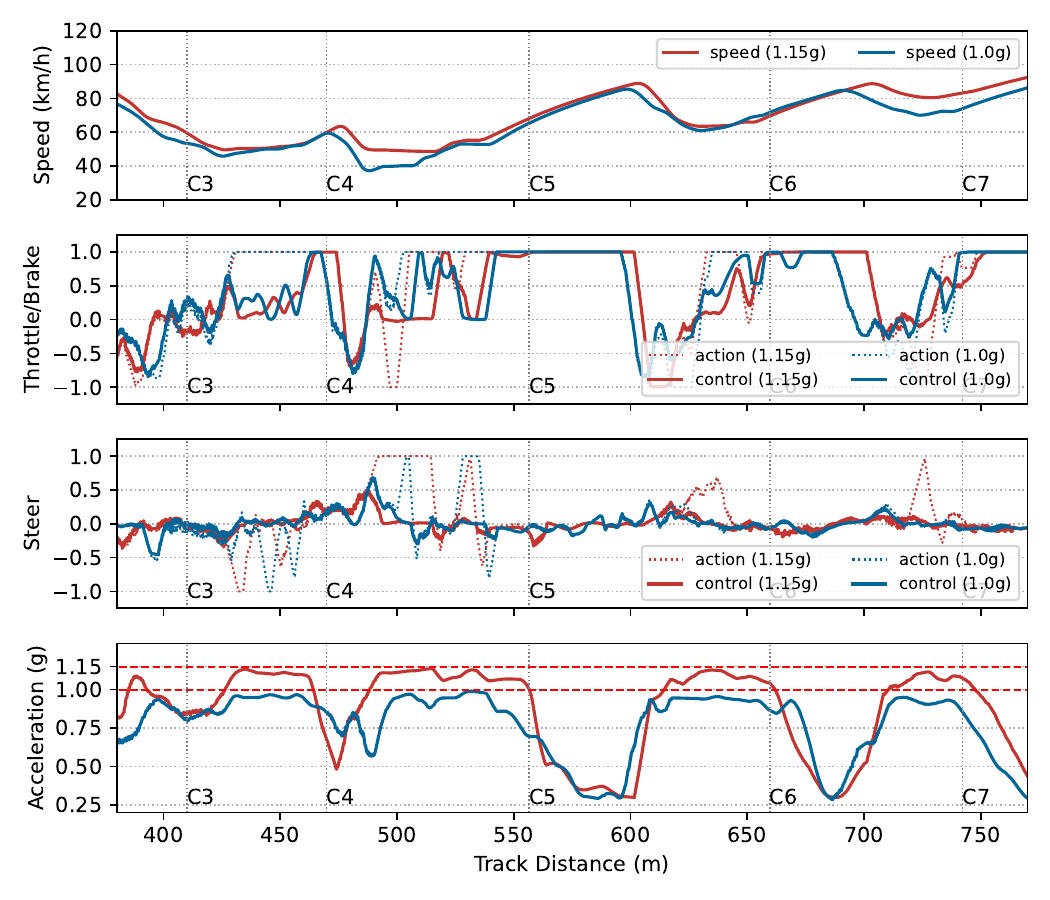}
  \caption{ Speed, throttle/brake action and control, steering action and control, and resultant acceleration by TD3-AM driving policy with different friction constraint ($1.15g$ and $1.0g$) in track-B corners C3 to C7. }
  \label{fig_newtrackstate2}
\end{figure}

In real race situations, the constraint of friction frequently changes due to tire wear, tire replacement, or wet track. If the maximum friction decreases while the original driving policy is in use, there is a higher risk of violating the friction constraint, particularly in sharp corners. The proposed TD3-AM driving policy can easily adapt to different friction constraints by adjusting the friction limit in the action mapping function. To demonstrate this feature, we compare the TD3-AM driving policy using two action mapping functions where the friction limits $\mu_{\max}$ are $1.15g$ and $1.0g$ respectively. We use the most difficult part of track-B, consecutive corners C3-C7, to demonstrate the performances. The trajectories are compared in Fig. \ref{fig_newtracktraj2}. The network output action $a_t$, control signal $u_t$ of throttle/brake and steering, and resultant acceleration are compared in Fig. \ref{fig_newtrackstate2} where the exit point of corners C3-C7 are marked. The trajectories of the driving policies with the constraint of $1.15g$ and $1.0g$ are similar, yet the speed of the policy with $1.0g$ policy is obviously slower. Since the maximum friction is lower, driving through a corner at a slower speed is the only way to avoid losing grip. 

From the resultant acceleration curves in Fig. \ref{fig_newtrackstate2}, both driving policies satisfy their corresponding constraint. The acceleration curves are very close to the upper limit in the corners, which means the policies could make the most of the tire grip to pass the corners at high speeds. That ability is mostly attributed to the AM mechanism. Specifically, from Fig. \ref{fig_newtrackstate2}, when the resultant acceleration approaches the limit in sharp corners, the constrained control input space becomes smaller. At this stage, whenever the network outputs an action that is outside the constrained space, the action mapping function gives the corresponding control signal that barely complies with the constraint. For example, in curves C3 and C4, the actor networks of both policies output the full-throttle action, while the acceleration curves indicate that turning at the current speed is very close to the friction limit. Then, the action mapping function gives a lower throttle signal to maintain the maximum possible speed through the corner. It should be noted that directly using an action mapping function with a lower friction limit can only guarantee that the driving policy satisfies the new friction constraint. The new driving policy cannot achieve the near-optimal driving performance as the original one, and it may not successfully finish a lap if the new friction limit is much lower. However, the new driving policy can be quickly improved and reach the near-optimal level with a few episodes of training in the lower friction situation. In this way, with the AM mechanism, the learned driving policy can quickly generalize to lower friction conditions without re-training the whole driving policy. 

Track-B is modeled after a real race track, and many cars have been tested on this track by professional drivers. Although the driving policy evaluation on track-B is performed in the simulation environment, we can make a rough comparison between our learned driving policy and the real professional drivers. We use the flying lap time data from \cite{kbracer}, and select six cars with similar lap times for comparison. The cars' basic performance parameters and lap times are listed in Table \ref{tb_laptime}. The maximum power (Max Pow.) and 0 to 100 km/h acceleration time (Acc. Time) indicate the car's acceleration performance; the 100 km/h to 0 brake distance (Brake Dist.) is associated with the maximum friction force. From these data, although our test model has little merit in basic performance parameters, the learned driving policy still managed to achieve a comparable lap time. Given the disparities between the simulation environment and the real-world scenario, we cannot directly compare the performances between the learned policy and the professional driver. However, the proposed TD3-AM algorithm for race driving has demonstrated its potential to acquire professional-level driving skills.  

\begin{table}[!ht]
\centering
\caption{Flying Lap Time Comparison on Ruisi Circuit (track B)}
\begin{tabular}{ccccc}
\toprule
Make/Model & \makecell[c]{Max Power \\ (kW)}  & \makecell[c]{Acc. Time \\ (s)} & \makecell[c]{Brake Dist \\ (m)} & \makecell[c]{Lap Time \\ (min:sec)} \\
\midrule
 BMW 325Li (2021)   & 135  & 7.90  & 38.7  & 1:03.28 \\
 Honda Civic (2022) & 134  & 8.13  & 38.2  & 1:03.91 \\
 VW Golf 8 (2021)   & 110  & 8.08  & 35.9  & 1:04.12 \\
 \rowcolor[gray]{.9} Our Test Model     & 125  & 8.80  & 42.5  & 1:04.16 \\
 Ford Focus (2019)  & 135  & 8.90  & 37.1  & 1:04.30 \\
 KIA K5 (2020)      & 176  & 7.48  & 34.1  & 1:05.10 \\ 
 Mazda Atenza (2020)& 141  & 8.03  & 37.0  & 1:05.40 \\
\bottomrule
\label{tb_trackb}
\end{tabular}

\label{tb_laptime}
\end{table}

\section{Conclusions and Future Work} \label{sec_conclusion}

 In this paper, we present a novel numerical AM-RL framework for autonomous race driving. The proposed numerical AM mechanism enables the RL-based driving agent to safely operate the vehicle within the friction limit while maximizing its handling capability. Leveraging the proposed TD3-AM approach, we have successfully trained a race driving agent with professional-level skills. The simulation results highlight improved race driving performance and the generalization capability to different friction conditions of the proposed approach, representing a significant advancement in addressing the friction constraint in autonomous racing. The lap time of the TD3-AM driving policy is 22\% shorter than the baseline TD3 driving policy, and the success rate is 90\%, which is much higher than the baseline policies.

In our future work, we aim to assess the practical applicability of our AM-RL framework through real-world car race experiments. Due to the unavailability of an accurate dynamic model for the vehicle, we intend to first establish a conservative friction constraint function using approximate physical parameters, and then enhance its accuracy progressively through online learning. We will also explore methods to handle sensor noise and other uncertainties in real-world applications. Additionally, we also plan to investigate the capabilities of AM-RL in addressing other nonlinear control problems with constraints.

\section*{References}
\bibliography{carrace_ref.bib}

\end{document}